\newif\ifcompsocstyle
\compsocstyletrue 
\ifcompsocstyle
\documentclass[10pt,journal,compsoc]{IEEEtran}
\else
\documentclass[lettersize,journal]{IEEEtran}
\fi

\ifCLASSOPTIONcompsoc
\usepackage[nocompress]{cite}
\else
\usepackage{cite}
\fi

\ifCLASSOPTIONcompsoc
\usepackage[caption=false,font=footnotesize,labelfont=sf,textfont=sf]{subfig}
\else
\usepackage[caption=false,font=normalsize,labelfont=sf,textfont=sf]{subfig}
\fi
\usepackage{ragged2e} 
\usepackage[top=0.4in,bottom=0.4in,left=0.4in,textwidth=7.7in]{geometry}
\usepackage[justification=justified,singlelinecheck=false]{caption}
\usepackage{amsmath,amsfonts}
\usepackage{algorithmic}
\usepackage{array}
\usepackage{textcomp}
\usepackage{stfloats}
\usepackage{url}
\usepackage{verbatim}

\usepackage[linesnumbered,ruled,vlined]{algorithm2e}  
\usepackage{enumitem}
\usepackage{graphicx}
\usepackage[colorlinks=true, linkcolor=blue, citecolor=blue]{hyperref}
\usepackage{float}
\usepackage{amssymb}
\usepackage{makecell}
\usepackage{booktabs} 
\usepackage{multirow}
\usepackage{etoolbox}
\usepackage[table]{xcolor}
\usepackage{xcolor}
\definecolor{DarkYellow}{rgb}{0.85,0.65,0.13}

\usepackage{pifont}

\usepackage{etoolbox}
\everydisplay{\small}

\makeatletter
\g@addto@macro\normalsize{%
  \setlength{\abovedisplayskip}{4pt}
  \setlength{\belowdisplayskip}{4pt}
  \setlength{\abovedisplayshortskip}{2pt}
  \setlength{\belowdisplayshortskip}{2pt}
}
\makeatother
\apptocmd{\normalsize}{\setlength{\abovedisplayskip}{4pt}%
                       \setlength{\belowdisplayskip}{4pt}}{}{}

\usepackage{multirow}
\NewDocumentCommand{\figref}{m o}{%
	Fig.~\hyperref[#1]{\ref*{#1}\IfValueT{#2}{(#2)}}%
}
\def\BibTeX{{\rm B\kern-.05em{\sc i\kern-.025em b}\kern-.08em
    T\kern-.1667em\lower.7ex\hbox{E}\kern-.125emX}}
\usepackage{balance}
\usepackage{setspace} 
\begin{document}
\bstctlcite{BSTcontrol}
\title{FUSAR-KLIP: Towards Multimodal Foundation Models for Remote Sensing}
\author{Yi Yang, Xiaokun Zhang, Qingchen Fang, Jing Liu, Ziqi Ye, Rui Li, Li Liu$^{\ast}$, Haipeng Wang$^{\ast}$

	\thanks{ \textit{}
		
		Yi Yang, Xiaokun Zhang, Qingchen Fang, Ziqi Ye, Rui Li, and Haipeng Wang are with the Key Laboratory for Information Science of Electromagnetic Waves (MoE), Fudan University, Shanghai 200433, China. Jing Liu is with the Institute of Zhejiang Laboratory, Hangzhou 310000, China. Li Liu is with the College of Electronic Science and Technology, NUDT, Changsha 410073, China. Corresponding authors are Li Liu and Haipeng Wang.

	}
}

\markboth{submitted to PAMI}%
{How to Use the IEEEtran \LaTeX \ Templates}

\IEEEtitleabstractindextext{%
\begin{abstract}
\justifying 
Cross-modal artificial intelligence, represented by visual language models, has achieved significant success in general image understanding. However, a fundamental cognitive inconsistency exists between general visual representation and remote sensing image interpretation: remote sensing images couple topography, terrain, and spatial structure, thereby inherently requiring models to possess deep geoscientific understanding. This cognitive difference is further amplified in synthetic aperture radar (SAR) imagery: while SAR possesses irreplaceable all-weather, all-day observation capabilities, it is constrained by coherent imaging mechanisms, exhibiting significant modal heterogeneity with general images. To address this inconsistency, we propose FUSAR-KLIP, the first knowledge-guided general multimodal foundational model for SAR, along with reusable data and evaluation baselines. Specifically: (1) FUSAR-GEOVL-1M (the first large-scale SAR dataset with complete geographic projection attributes) was constructed, covering multiple satellite platforms, 120,000 images, and 135 cities; (2) Aligned structured text was generated through hierarchical cognitive thought chains, accurately encoding more than 1 million multidimensional semantic information from geomorphological environment and regional attributes to spatial relationships; (3) A self-consistent iterative optimization mechanism was designed to guide cross-modal learning with this knowledge information consistent with human cognition and physical laws in a self-supervised closed loop consisting of contrast, matching, and reconstruction; (4) A unified evaluation benchmark was established in 11 typical downstream tasks in the two major categories of vision and language, and compared with 15 mainstream foundation models. Experiments show that FUSAR-KLIP exhibits optimal performance, paving a new path for building remote sensing intelligent systems that are more in line with human cognitive logic. The dataset and model are publicly available at: \url{https://github.com/yangyifremad/FUSAR-KLIP}.
\end{abstract}

}
\maketitle
\IEEEdisplaynontitleabstractindextext

\section{Introduction}\label{sec1}

\IEEEPARstart{I}{n} recent years, foundational models built around the Transformer have established a dominant position in the field of computer vision, demonstrating outstanding general representation capabilities through self-supervised paradigms such as masked image modeling or contrastive learning\cite{10490262}. Particularly, the advent of vision-language models have pioneered a new paradigm of cross-modal semantic alignment, enabling models to transcend simple visual feature extraction and achieve high-level cognitive understanding\cite{11184436}. Inspired by this, a number of foundational models targeting the characteristics of Earth observation have rapidly emerged in the field of remote sensing. To address the need for high-precision sensing, RingMo\cite{diao2025ringmo} and SpectralGPT\cite{10490262} have achieved deep adaptation to small targets and multispectral features through customized masked modeling and 3D spectral reconstruction. Meanwhile, in order to bridge the gap between low-level pixels and high-level semantics, GeoRSClip\cite{10679571} and RemoteCLIP\cite{10504785} introduce natural language alignment to enhance the inference depth of the model. To address the challenges of wide-area generation and data sparsity, MetaEarth\cite{yu2024metaearth} and AlphaEarth\cite{brown2025alphaearth} have introduced generative diffusion and spatiotemporal embedding field techniques, successfully extending the model's capabilities from passive interpretation to active simulation and multi-source assimilation\cite{11267238},\cite{10752552}, forming a diversified development pattern centered around the core needs of Earth observation\cite{10607962},\cite{zhou2024vlgfm},\cite{10770756},\cite{soni2025earthdial},\cite{liuRandomFeatures}.

While the aforementioned work has achieved record-breaking performance metrics on specific tasks, its theoretical foundation still faces significant challenges. As Muttenthaler et al. pointed out, a fundamental misalignment exists between intelligent model representations and human cognition: existing deep neural networks, although excelling in local feature matching, often fail to capture hierarchical knowledge structures like humans\cite{muttenthaler2025aligning}. This means that current models lack the deep cognitive abilities that align with human logic, resulting in significantly limited robustness and generalization when faced with out-of-distribution data\cite{9638340}.

\begin{figure}[!tbp]
	\centerline{\includegraphics[width=0.5\textwidth]{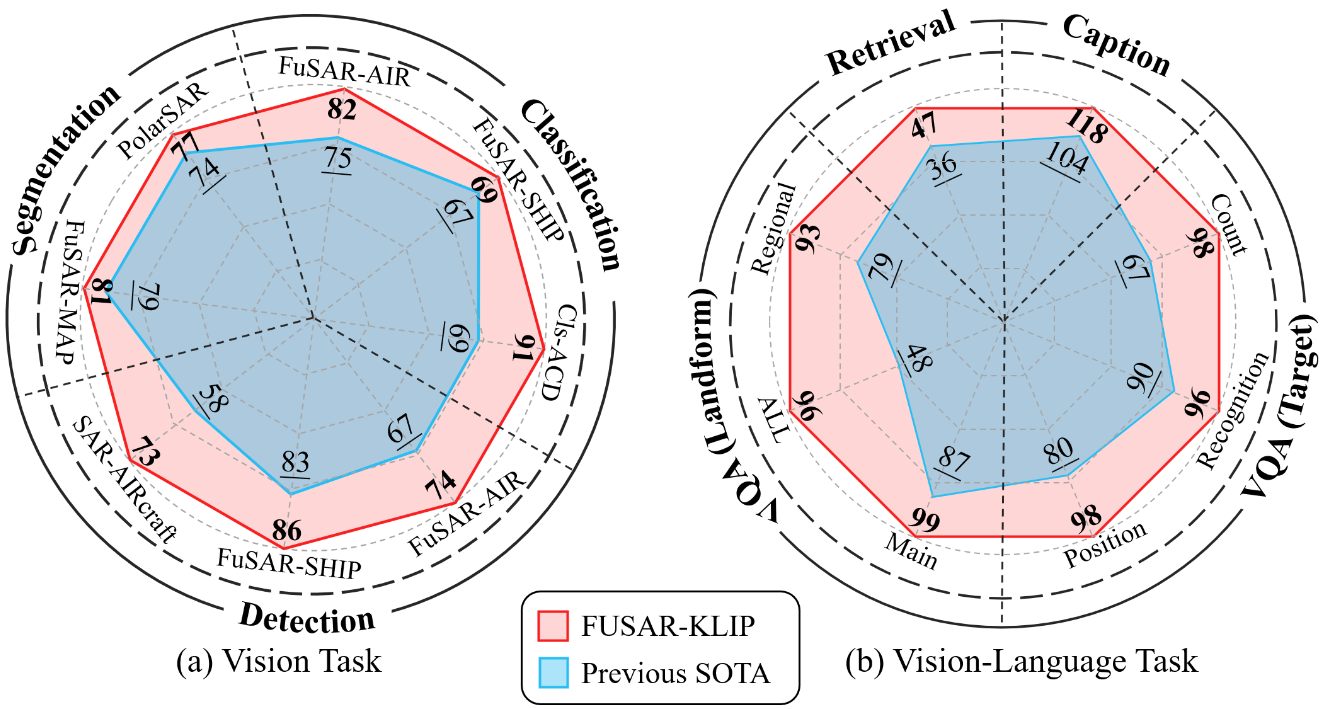}}
	\caption{Performance of the foundation models on SAR interpretation tasks after fine-tuning. Previous SOTA represents the performance of the best model among the compared methods for this task.}
	\label{radar}
\end{figure}

This cognitive misalignment is particularly acute in the field of Earth observation. Because it hasn't fundamentally broken free from the established framework of "general vision," existing remote sensing models tend to treat images as high-resolution natural pictures for visual understanding, neglecting the unique characteristics of remote sensing interpretation logic: remote sensing images are not merely collections of pixels, but deeply coupled entities of geomorphic features, regional attributes, and spatial structures. Lacking the ability to model hierarchical geographic information, these models often remain at a superficial level of visual perception, failing to possess deep geoscientific cognitive capabilities.

This cognitive disparity has further evolved into a dual barrier of modal heterogeneity and mechanism intricacy in the field of synthetic aperture radar (SAR)\cite{11079686}. Although SAR possesses irreplaceable all-weather strategic value, its imaging, based on microwave coherent mechanisms, is heavily affected by speckle noise and geometric distortion\cite{10638186}. When existing foundation models are migrated to the SAR domain, they struggle to understand the geometric structure and physical properties of ground objects through non-intuitive scattering phenomena, leading to a sharp decline in generalization performance\cite{zhou2025fifty}. Therefore, leveraging the advantages of cross-modal artificial intelligence to enhance cognitive capabilities through vision-language models and constructing a multimodal foundation model that can adapt to SAR imaging characteristics and possess geospatial cognitive capabilities has become a key scientific task in the field of remote sensing intelligence.

Realizing this vision is no easy task. Due to limitations in data privacy, domain specialization, and feature complexity, this research direction still faces many challenges, specifically:

\textit{1) \textbf{Geospatial Priors Remain Underutilized:}} Geographic information, a core element in Earth sciences, is crucial for remote sensing. However, most current SAR interpretation studies remain at the visual level, neglecting geographic attributes. This stems from the extension of natural image paradigms—focused mainly on texture and structure\cite{Liu2016MRELBP}—as well as the closed nature of SAR data sources, with many studies still relying on outdated datasets like MSTAR from the 1990s\cite{AFRL_MSTAR}. As shown in Table \ref{tab1}, our review of 16 mainstream SAR datasets across tasks such as detection, segmentation, and recognition reveals a widespread omission of geographic metadata. This absence hinders regional analysis, terrain understanding, and spatial reasoning, ultimately limiting progress toward cognitive-level SAR interpretation.

\begin{figure}[!tbp]
	\centerline{\includegraphics[width=0.45\textwidth]{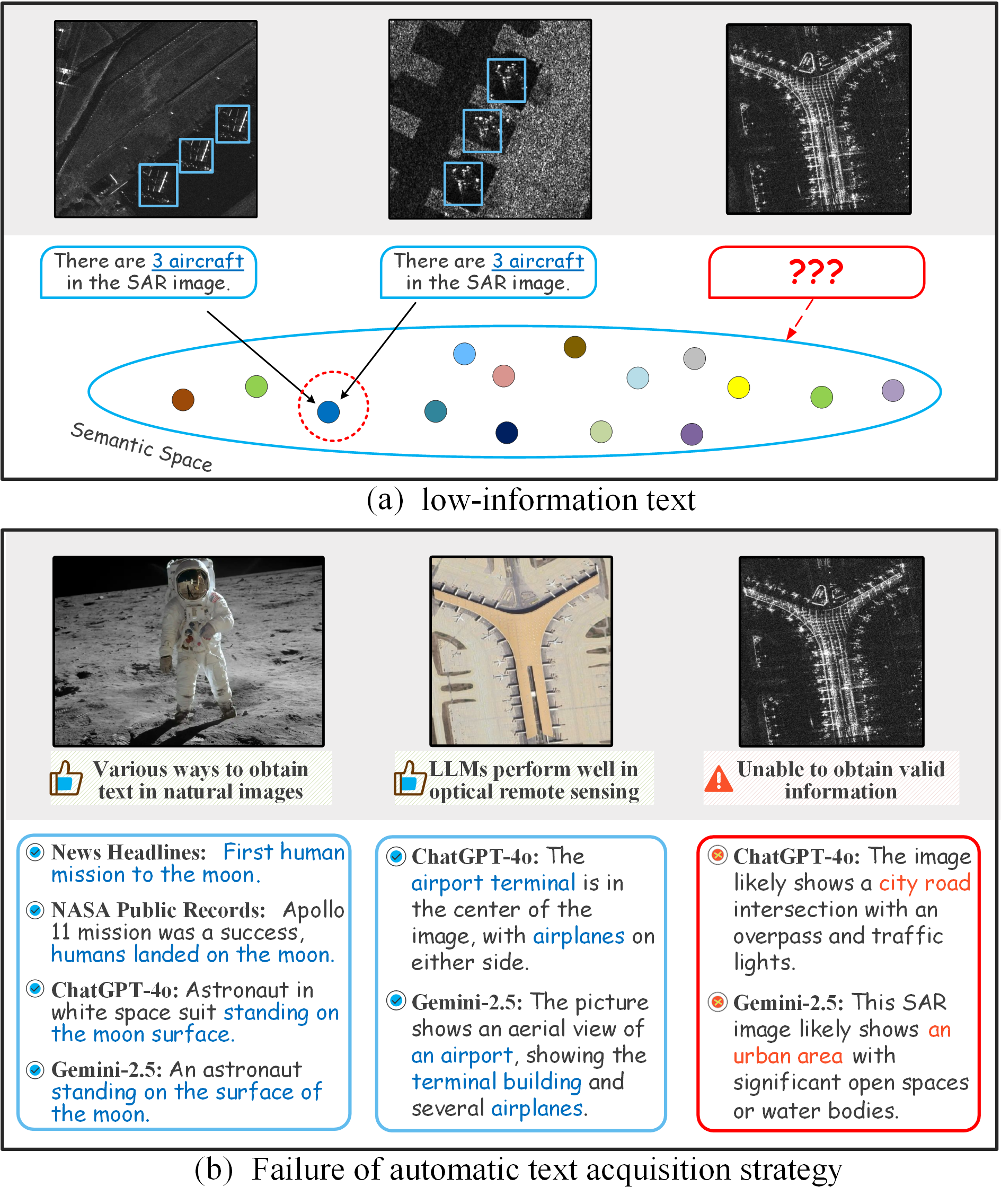}}
	\caption{The construction of text data for SAR images encounters a knowledge bottleneck.}
	\label{fig1}
\end{figure}

\textit{2) \textbf{Knowledge Bottleneck:}} Existing remote sensing VLM studies face a significant knowledge bottleneck in text construction. On the one hand, approaches that convert detection, recognition, or segmentation annotations into templated text \cite{10504785},\cite{wei2025sarlang} generate descriptions that are semantically sparse and structurally repetitive, covering only shallow attributes (e.g., category and location) while neglecting complex elements such as landform, regional function, or spatial structure. This low-information text causes different images to collapse into similar semantic spaces, weakening alignment quality and fine-grained discrimination, as shown in \figref{fig1}[a]. On the other hand, automatic text acquisition strategies, which are effective in natural image or optical remote sensing domains (e.g., web crawling, video subtitles, or AI-generated captions), fail in the SAR domain due to speckle noise and the need for expert knowledge. The resulting AI-generated text is often semantically vague or distorted \figref{fig1}[b]. Together, these issues highlight that SAR multimodal foundational model suffers from a dual bottleneck: text derived from manual annotations lacks semantic depth, and reliable large-scale automatic acquisition remains infeasible, severely restricting scalability in open-world applications.

\textit{3) \textbf{Inadequate cross-modal alignment and optimization:}} Current multimodal methods largely depend on direct image–text alignment, without deep feature fusion or closed-loop optimization, which constrains adequacy and robustness. Their performance is also susceptible to text quality—an acute issue in remote sensing, where image descriptions often contain bias and automatically generated text is prone to semantic inaccuracies. As a result, existing approaches remain limited in achieving reliable semantic completion and fine-grained understanding.

\begin{table}[!t]

	\caption{Current Status of SAR Public Datasets: Lack of Geographic Information. Pol: Polarization. Res: Resolution. Geo: Geographic Information. Cls: Classification. Det: Detection. Seg: Segmentation. Cap: Caption. VQA: Visual question answering. Ret: Retrieval}
	\renewcommand{\arraystretch}{1.5}
	\centering

	\label{tab1}
	\setlength{\tabcolsep}{3pt}
	\resizebox{0.5\textwidth}{!}{  
		\begin{tabular}{l c c c c c c c c c c c}  
			\hline
			\textbf{Dataset} & \textbf{Year} & \textbf{Cite} & \textbf{Band} & \textbf{Pol} & \textbf{Res} & \textbf{Size} & \textbf{Quantity} & \textbf{Format} & \textbf{Task} & \textbf{Text} & \textbf{Geo} \\
			\hline
			MSTAR\cite{AFRL_MSTAR} & 1995 & 1155 & X & Single & 0.3 & 128$\sim$193 & 14,577 & JPG &  Classification & \ding{55} & \ding{55} \\
			OpenSARShip\cite{8067489} & 2018 & 299 & C & Multi & 10 & 30$\sim$120 & 11,346 & TIF &  Classification & \ding{55} & \ding{55} \\
			FUSAR-MAP\cite{shi2021object} & 2021 & 45 & C & Single & 1 & 1,024 & 610 & TIF & Segmentation & \ding{55} & \ding{55} \\
			SSDD\cite{XU2025440} & 2021 & 633 & C,X & Multi & 1$\sim$15 & 500 & 1,160 & JPG & Detection & \ding{55} & \ding{55} \\
			MSAR\cite{10057265} & 2022 & 23 & X & Single & 1 & 256$\sim$2,048 & 8,449 & JPG & Detection & \ding{55} & \ding{55} \\
			SADD\cite{9761751} & 2022 & 75 & X & Single & 0.5$\sim$3 & 224 & 2,966 & BMP & Detection & \ding{55} & \ding{55} \\
			SAR-Ship\cite{8878083} & 2019 & 464 & C & Multi & 3$\sim$22 & 256 & 43,819 & TIFF & Detection & \ding{55} & \ding{55} \\
			SAR-ACD\cite{9754573} & 2022 & 55 & C & Single & 1 & 32$\sim$200 & 3,032 & JPG & Detection & \ding{55} & \ding{55} \\
			FUSAR-SHIP\cite{hou2020fusar} & 2022 & 271 & C & Single & 1 & 512 & 16,144 & TIFF & Detection & \ding{55} & \ding{55} \\
			SARDet-100k*\cite{li2024sardet100k} & 2024 & 48 & Mix & Multi & 0.1$\sim$1.5 & 512 & 116,598 & PNG & Detection & \ding{55} & \ding{55} \\
			SAMPLE\cite{lewis2019sar} & 2019 & 14 & X & Single & 0.3 & 128 & 2,732 & PNG & Classification & \ding{55} & \ding{55} \\
			AIR-SARShip\cite{Sun2019AIRSARShip} & 2019 & 248 & C & Single & 1$\sim$3 & 1,000 & 300 & TIFF & Detection & \ding{55} & \ding{55} \\
			SAR-AIRcraft\cite{Wang2023SARAIRcraft} & 2023 & 101 & C & Single & 1 & 800$\sim$1,500 & 4,368 & JPG & Detection & \ding{55} & \ding{55} \\
			PolarSAR-Seg\cite{wang2022air} & 2022 & 36 & C & Multi & 8 & 512 & 500 & TIFF & Segmentation & \ding{55} & \ding{55} \\
			SARATR-X*\cite{10856784} & 2025 & 10 & Mix & Multi & 0.1$\sim$1.5 & mix & 180,000 & PNG & Detection & \ding{55} & \ding{55} \\
			ATRNet-STAR\cite{liu2025atrnet} & 2025 & - & X,Ku & Multi & 0.12$\sim$0.15 & 128 & 194,324 & TIF & Classification & \ding{55} & \ding{55}\vspace{3pt} \\
			\hline 
			\rowcolor{lightgray}\textbf{FUSAR-GEOVL-1M} & \textbf{2025} & \textbf{-} & \textbf{C,X,Ku} & \textbf{Multi} & \textbf{0.5$\sim$3} & \textbf{256$\sim$5,120} & \textbf{120,000} & \textbf{TIFF} & \makecell[c]{\textbf{Cls,Det,Seg} \\ \textbf{VQA,Cap,Ret} } & \textbf{\ding{51}} & \textbf{\ding{51}} \\
			
			\hline
			\multicolumn{12}{l}{\textit{Note:} Datasets marked with * are collected from multiple previously released datasets.} \\
		\end{tabular}
	}
\end{table}

To address these challenges, we present FUSAR-GEOVL-1M, the first large-scale SAR image-text dataset enriched with open text description and geographic metadata, along with FUSAR-KLIP, the first universal multimodal foundational model for SAR images. \textbf{FUSAR-KLIP} is a \textbf{K}nowledge-guided \textbf{L}anguage-Image \textbf{P}re-training model that aims to move beyond the limitations of single-modality visual perception, establishing a multimodal \textbf{SAR} foundation model with cognitive reasoning, semantic representation, and transferable capabilities.

\textbf{In terms of image data construction}, FUSAR-GEOVL-1M contains 120,000 images from three SAR satellite platforms, spanning multiple resolutions, 135 cities, and five landform types. To mitigate semantic inconsistency caused by cross-platform resolution variation, we introduce a Spatial Resolution Consistency (SRC) slicing strategy, which aligns semantic granularity at the geographic level and ensures uniform cognitive scale for model training.

\textbf{In terms of text data acquisition}, this paper introduces a hierarchical cognitive chain-of-thought (HCoT) instruction to simulate the human interpretation process of SAR imagery. HCoT guides GPT-4.1 \cite{OpenAIChatGPT} to progressively incorporate multi-dimensional knowledge—such as geographic context, regional priors, SAR imaging principles, and target scale perception—enabling semantic reasoning from global to local, and from general to domain-specific knowledge. Furthermore, the constructed multi-scale image semantics driving mechanism guides the model to generate semantic expressions that connect the upper and lower scales on large, medium, and small scale images, and establish cognitive coherence.  Under HCoT guidance, the language model significantly enhances its SAR interpretation ability, generating more informative and coherent textual descriptions to support multimodal representation learning.

\textbf{In terms of multimodal alignment modeling}, this paper proposes FUSAR-KLIP, a dual-tower multimodal framework employing Vision Transformer (ViT)\cite{dosovitskiy2021an} and BERT\cite{devlin2019bert} as visual and language encoders. The model jointly optimizes image-text contrastive loss (ITC), image-text matching loss (ITM), and masked language modeling loss (MLM) to construct a unified cross-modal embedding space bridging low-level perception and high-level semantics. To mitigate semantic deviations in HCoT-guided generated text, we introduce a self-consistent iterative optimization (SCIO) module that enhances alignment accuracy and stability through a closed-loop self-supervised strategy of screening, filtering, and reconstruction.

The main contributions of this paper are summarized as follows:
\begin{itemize}[leftmargin=1em, itemsep=2pt]
    \item \textbf{FUSAR-GEOVL-1M Dataset:} This dataset represents the first large-scale SAR image and text dataset with complete geographic information. It encompasses data from three types of SAR satellite platforms, 135 cities, and multi-scale typical scenes, including over 120,000 images and more than one million text descriptions. By addressing the absence of geographic attributes in SAR image interpretation research, this dataset fills a critical gap and provides a foundational data resource for SAR multimodal modeling research.
    \item \textbf{Text Generation Mechanism Guided by HCoT:} The paper designs the HCoT instruction system to simulate the human reasoning process, guiding the large language model to integrate multi-dimensional knowledge and generate structured semantic information. This approach establishes a new paradigm for SAR image text annotation that is independent of manual intervention, explainable, and scalable.
    \item \textbf{FUSAR-KLIP (Knowledge-guided Language-Image Pre-training):} The first knowledge-guided visual language foundation model for SAR was constructed, combining contrast, matching, and reconstruction multi-task learning to establish a cross-modal representation space for vision-language collaboration. The SCIO optimization module is introduced to dynamically enhance text accuracy and improve cross-modal alignment quality through the "screen–filter–reconstruct" self-supervised closed-loop optimization mechanism.
    \item \textbf{Leading Multi-Task Generalization Capability:} In various typical downstream tasks, such as target classification, detection, land feature segmentation, image captioning, image-text retrieval, and visual question answering, FUSAR-KLIP demonstrates superior semantic understanding and cross-task generalization performance compared to existing remote sensing multimodal models.

\end{itemize}

The remainder of this paper is organized as follows: Section \ref{sec2} introduces the research progress in related fields; Section \ref{sec3} details the construction process of the FUSAR-GEOVL-1M dataset and the design of the multimodal self-supervised model; Section \ref{sec4} presents experimental verification and analysis across multiple remote sensing downstream tasks; Section \ref{sec5} provides a summary of the paper and outlines potential future research directions.

\section{Related works}\label{sec2}

\subsection{Paradigm Evolution of Remote Sensing Interpretation}

\begin{figure}[!bp]
	\centerline{
		\phantomsection
		\includegraphics[width=0.47\textwidth]{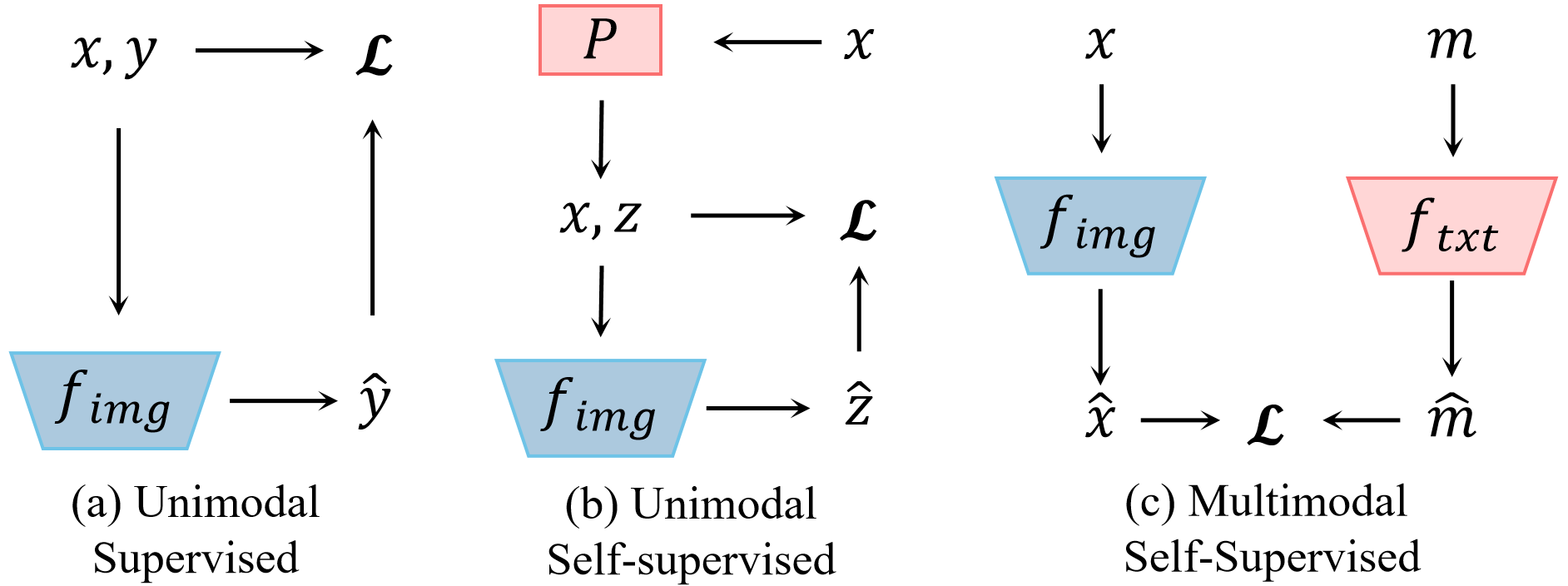}}
	
	\caption{The evolution of remote sensing interpretation research: from unimodal supervised learning to multimodal self-supervised learning.}
	\label{fig2}
\end{figure}

As illustrated in \figref{fig2}, the research paradigm for intelligent remote sensing interpretation has evolved from supervised learning to unimodal self-supervision and, more recently, to multimodal self-supervision. 

\textbf{In the supervised learning stage,} large-scale manual annotations enabled significant progress in recognition, detection, and segmentation tasks, but the prohibitive cost and limited semantic coverage of annotations severely constrained generalization across diverse scenes\cite{shi2021object}.

\textbf{Unimodal self-supervised learning (SSL)} alleviated this bottleneck by mining latent structures from unlabeled data, producing robust representations with minimal fine-tuning and laying the foundation for domain-specific models\cite{10713915},\cite{10955443}. The RingMo series pioneered the introduction of mask learning into remote sensing, optimizing feature extraction of dense small targets in remote sensing images through targeted masking strategies\cite{diao2025ringmo}. RingMo-Aerial further introduced a frequency-enhanced attention mechanism and affine transformation contrastive learning, solving the challenges of tilted viewpoints and multi-scale occlusion in aerial remote sensing. Addressing the 3D characteristics of multispectral data, SpectralGPT innovatively employs 3D masking and multi-target reconstruction strategies, overcoming the limitation of processing only RGB data and achieving effective capture of spectral sequence information\cite{10490262}. In the generative direction, MetaEarth, through a self-cascaded framework and noise sampling strategy, achieved global-scale, unbounded resolution remote sensing image generation, driving the leap from perception to simulation\cite{yu2024metaearth}. RS-vHeat takes a different approach, using a heat conduction physics model to simulate feature diffusion, significantly improving computational efficiency while maintaining the global receptive field\cite{hu2025rs}. Furthermore, AlphaEarth proposed the concept of Embedding Fields, addressing the label sparsity problem by assimilating multi-source spatiotemporal data\cite{brown2025alphaearth}. Nevertheless, unimodal SSL remains largely restricted to low-level perceptual cues such as texture and geometry, offering limited capacity for higher-order semantic reasoning\cite{10915556},\cite{10955443}.

\textbf{Multimodal SSL} further introduces language to build cross-modal embedding spaces through large-scale image–text alignment, enabling richer semantics and task generalization. This paradigm has driven breakthroughs in natural imagery and optical remote sensing. However, SAR has received little attention within this framework: most existing models rely on RGB-based assumptions and fail to capture SAR’s unique imaging mechanisms. As a result, researchers have only begun to explore unimodal SSL in SAR as a transitional step toward multimodal expansion.

\subsection{Research on Unimodal Self-supervised Foundational Models in SAR Images}

In the field of SAR imagery, researchers have begun to incorporate the SSL paradigm into the modeling process to alleviate the scarcity of labeled data and improve the performance of downstream tasks. Existing work has explored the potential of unimodal SSL from different perspectives.

SARATR-X uses a two-step SSL method with multi-scale gradient features to establish a high-performance SAR image target recognition foundational model\cite{10856784}. Yang et al. proposed the SARDet-CL, combining feature enhancement with a SSL method constrained by imaging mechanisms, and achieved advanced performance in downstream detection tasks\cite{10994283}. Li et al. proposed SAR-JEPA, which constructed a self-supervised pre-training foundational model for SAR target recognition tasks from the perspective of overcoming speckle noise\cite{LI2024326}. Ren et al. introduced multi-image factor SSL to promote directional feature learning and obtain generalized features, enhancing the performance in terrain classification tasks\cite{10496083}. Wang et al. constructed a SAR multi-task foundation model based on cross-domain continuous pre-training\cite{10726860}. Pei et al. proposed a two-stage SAR image pre-training method based on SSL to improve the accuracy of target classification\cite{10269665}. MSFA proposed a multi-stage filter augmentation pre-training framework for large-scale RGB and SAR data, which performed well when transferred to detection tasks\cite{li2024sardet100k}.

While these methods demonstrate the potential of SSL for SAR representation, their effectiveness often relies on small- to medium-sized, single-task datasets, and their generalization across platforms, multi-polarization, and open scenarios has yet to be fully validated. Furthermore, they primarily focus on low-level semantic tasks such as classification and detection, making them incapable of supporting complex scene parsing and cross-regional understanding\cite{10630605}. More importantly, the learned representations remain limited to perceptual-level features such as texture and geometry, lacking the modeling and integration of linguistic semantics and geographic priors. Therefore, the exploration of unimodal SSL in SAR remains preliminary and needs to be expanded to multimodal modeling to achieve higher-level semantic cognition and task generalization\cite{10504785}.

\subsection{Research on Multimodal VLMs in Remote Sensing Field}
Driven by cross-modal learning, the remote sensing field has also begun exploring vision-language models, hoping to obtain transferable high-level semantic representations through image-text alignment. 

Existing work primarily focuses on optical imagery\cite{11095174}. For instance, RemoteCLIP is based on the CLIP framework by converting class labels in public remote sensing datasets into templated text, establishing a foundational model that achieves leading performance across many typical tasks\cite{10504785}. SkyEyeGPT developed a multimodal remote sensing command dataset and designed a two-stage tuning strategy to enhance conversational capabilities\cite{zhan2025skyeyegpt}. VHM generated the HqDC dataset based on the large language model Gemini, easing the issue of model hallucination\cite{pang2025vhm}. SkyScript utilized structured geographic information from OpenStreetMap to generate semantic text for optical images, improving the quality of cross-modal alignment\cite{wang2024skyscript}. GeoChat expanded existing remote sensing image-text pairs to build a multi-round conversation dataset with command-following capabilities\cite{kuckreja2024geochat}. BITA introduced a lightweight Fourier Transformer structure to enhance image-text interaction\cite{10415446}, while BAN developed a foundational model for remote sensing change detection tasks, leading to improved task performance\cite{10438490}. Multimodal modeling for SAR images is still in its infancy. 

In contrast, multimodal foundation model construction for SAR is still in its infancy: early attempts such as SARLANG\cite{wei2025sarlang} and SARCLIP\cite{jiang2026sarclip}, which generates text from detection annotations, and SSL-LIP\cite{10642118}, which employs dual-stage self-supervised training with limited labels, have demonstrated feasibility but remain constrained by small-scale, template-based text construction and sparse semantics. Therefore, there is an urgent need to construct a multimodal foundation model adapted to the imaging characteristics of SAR images.

\section{Method}\label{sec3}

\subsection{FUSAR-GEOVL-1M Image Data}\label{sec3.B}
\begin{figure*}[htbp]
	\centerline{
		\phantomsection
		\includegraphics[width=0.9\textwidth]{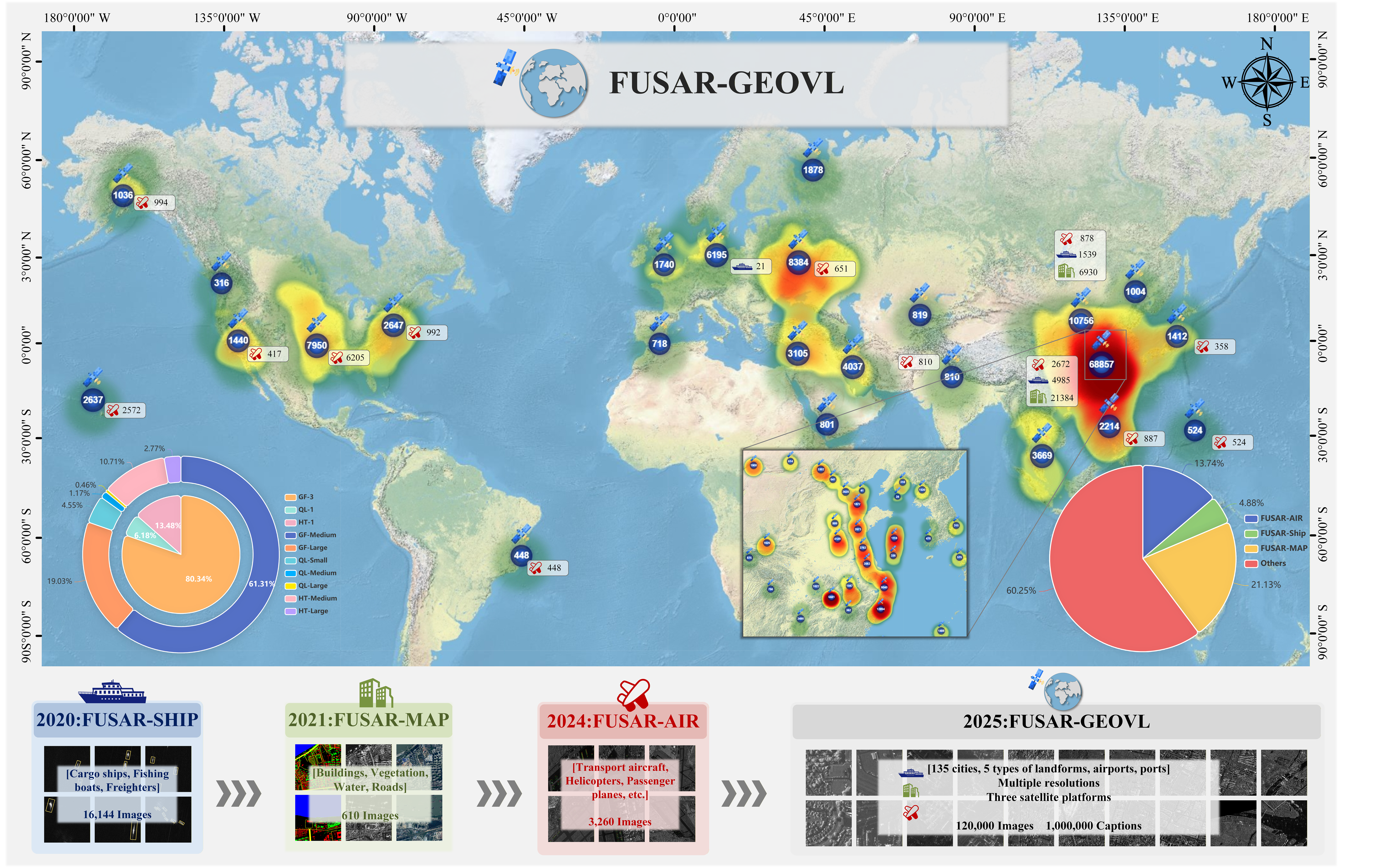}}
	
	\caption{FUSAR-GEOVL-1M image data. FUSAR-SHIP: A ship detection dataset we released in 2020. FUSAR-MAP: A land feature classification dataset we released in 2021. FUSAR-AIR: An aircraft detection dataset we released in 2024. FUSAR-GEOVL: A large-scale SAR multimodal dataset proposed in this study.}
	\label{fig4}
\end{figure*}
Given that existing SAR datasets generally lack geospatial attributes, making them difficult to support complex semantic modeling and spatial reasoning, this study's approach first addresses the data layer. Current mainstream datasets (as shown in Table \ref{tab1}) often lack geographic metadata, resulting in a disconnect between imagery and real-world spatial locations, limiting tasks such as regional functional reasoning and multi-scale target analysis. To this end, we constructed FUSAR-GEOVL-1M, the first large-scale SAR dataset to fully preserve geographic information. Its core design principle is spatial scale consistency: it fuses SAR imagery from multiple platforms and resolutions, and provides each image with WGS84 projection coordinates, enabling precise spatial positioning and adaptability to multimodal tasks.

The FUSAR-GEOVL-1M dataset is collected from three SAR satellites with different resolutions, including Qilu-1\cite{10948491}, Gaofen-3\cite{9585657}, and Hongtu-1\cite{10816656}. Qilu-1 has a resolution of 0.2 meters and operates in the Ku band, offering high-precision urban modeling capabilities. Gaofen-3 has a resolution of 1 meter and operates in the C band, which is suitable for regional target and structure extraction. Hongtu-1 has a resolution of 3 meters and operates in the X band, ideal for modeling urban area edges and terrain structures.

The dataset covers 135 representative urban areas, as shown in \figref{fig3}. The scene types include typical remote sensing semantic categories such as airports, ports, urban areas, water bodies, industrial parks, and road networks. The dataset construction process consists of the following key steps:

\begin{figure}[htbp]
	\centerline{
		\phantomsection
		\includegraphics[width=0.35\textwidth]{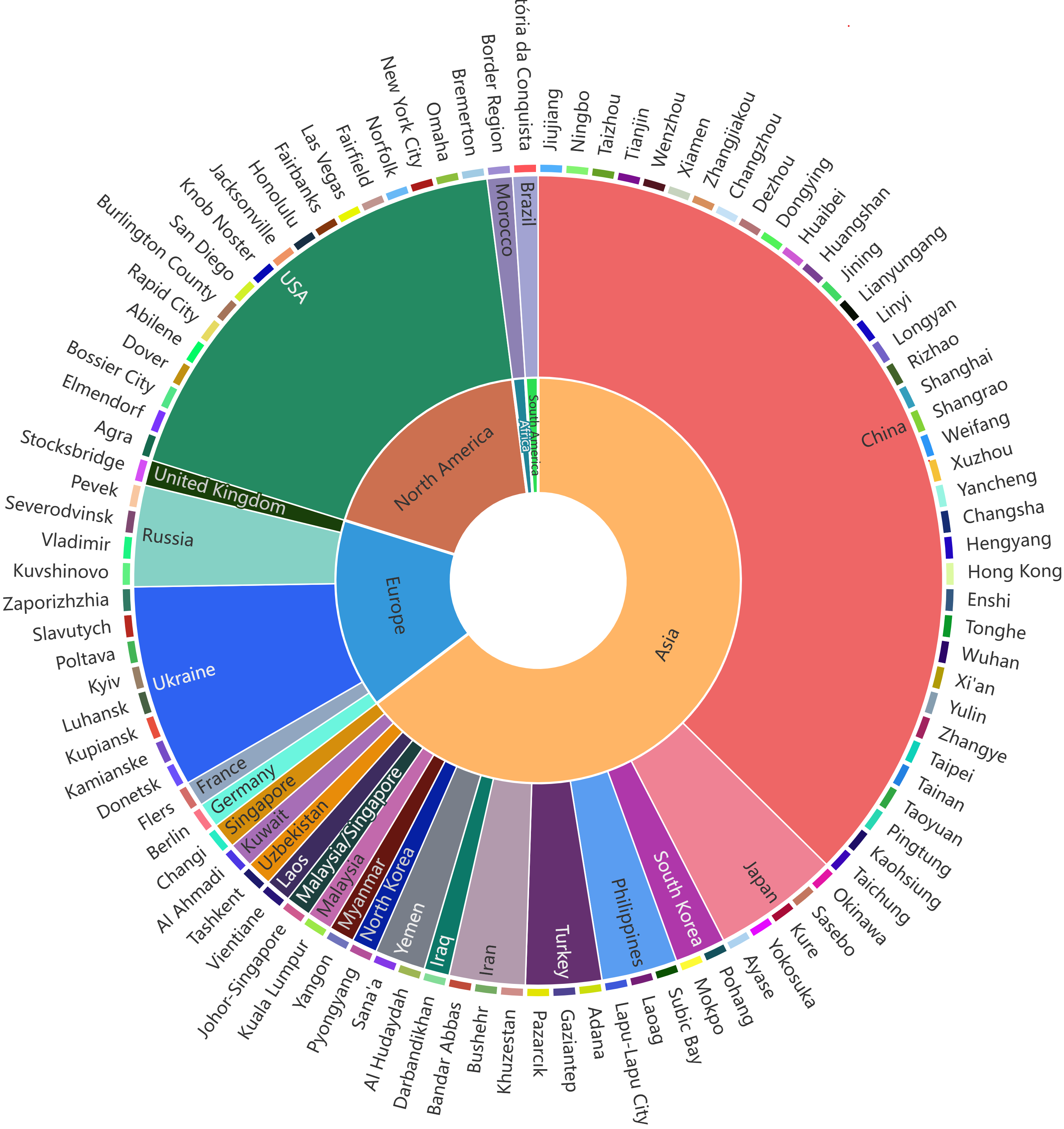}}
	
	\caption{Cities covered by the FUSAR-GEOVL-1M dataset.}
	\label{fig3}
\end{figure}

\textit{1) \textbf{Region Screening}:} High-semantic-density regions—such as airports, ports, and urban built-up areas—are selected from full-scene SAR imagery, while low-structure regions (e.g., grasslands, water bodies) are excluded to ensure diversity and relevance.

\textit{2) \textbf{SAR Image Preprocessing}:} Original 16/32-bit float SAR images undergo dynamic range compression and threshold quantization following SARDet-CL\cite{10994283}, standardizing them to 8-bit grayscale (uint8). This enhances feature clarity while reducing storage and I/O overhead.

\textit{3) \textbf{Spatial Scale Consistency Strategy (SRC)}:} Due to resolution disparities across SAR satellite platforms, fixed-size cropping may result in inconsistent ground coverage and semantic ambiguity. To resolve this, we propose a SRC strategy that adaptively adjusts the cropping window according to spatial resolution, ensuring each slice represents a uniform geographic area (e.g., 1m resolution images are cropped to 1024×1024 pixels, and 0.2m resolution images are cropped to 5120×5120 pixels, both covering 1 km²).

\textit{4) \textbf{Coordinate Mapping}:} Geographic coordinates are recalculated using affine transformation and WGS84 projection, enabling accurate spatial referencing across scales and time, and supporting spatiotemporal sequence tasks.

\textit{5) \textbf{Quality Screening and Filtering}:} Post-processing employs GLCM\cite{752194} and statistical features, with a KNN-based\cite{6409354} filter to remove low-information or structurally deficient samples, improving dataset quality.

As shown in \figref{fig4}, the FUSAR-GEOVL-1M dataset presents a clear distribution of data, sample proportions, and representative examples. It is noteworthy that this dataset is built upon several task-oriented SAR datasets previously proposed by our team—namely, FUSAR-MAP\cite{shi2021object}, FUSAR-SHIP\cite{hou2020fusar}, and FUSAR-AIR\cite{Qian2024Thesis}—which have undergone large-scale expansion and data restructuring. These enhancements have significantly improved the dataset’s coverage, semantic density, and adaptability to downstream tasks. The aforementioned datasets have been downloaded over 6,000 times and are widely cited and applied in the field of SAR interpretation. FUSAR-GEOVL-1M represents the systematic accumulation of our team’s long-term research in SAR image understanding and provides a strong data foundation for multimodal modeling in the SAR domain.

\begin{figure*}[htbp]
	\centerline{
		\phantomsection
		\includegraphics[width=0.9\textwidth]{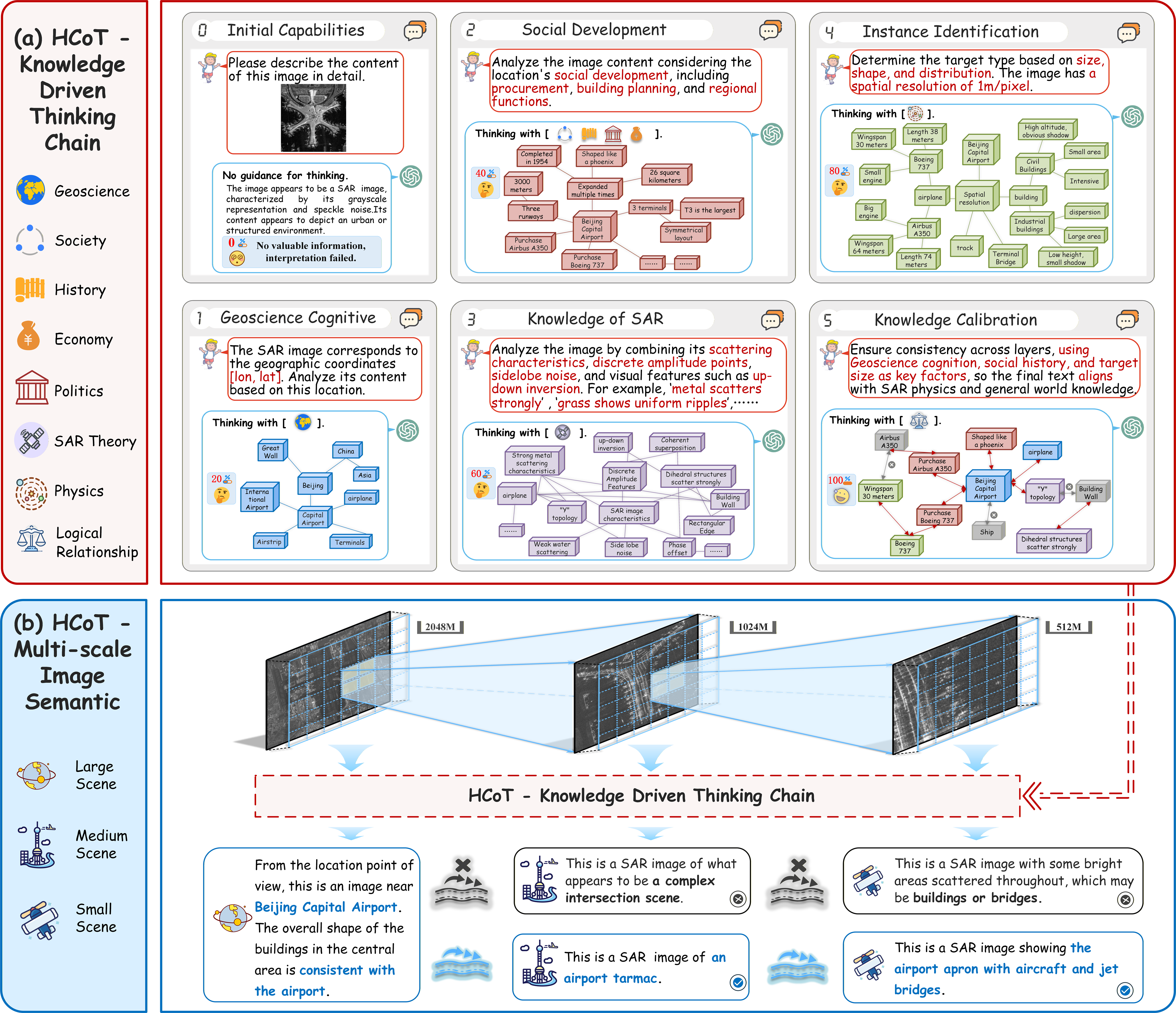}}
	
	\caption{(a) A knowledge-driven thinking chain prompt word system is established to achieve the automatic and effective acquisition of SAR image text information. (b) Multi-scale image semantics-driven thinking chain prompt. The thought chain prompt based on multi-scale image semantics enables the interpretation of remote sensing images to consider global and local complementary information.}
	\label{fig5}
\end{figure*}

\subsection{Text Generation Guided by Hierarchical Cognitive Chain-of-Thought (HCoT)}\label{sec3.c}
After constructing a SAR image dataset with spatial semantic consistency and multi-scale feature expression capabilities, a more core challenge arises: how to generate interpretable and structured language descriptions for each image to support cross-modal training. Although modern LLMs possess extensive general and domain knowledge and perform well on optical imagery tasks, they often fail to semantically interpret SAR content due to the unique electromagnetic imaging mechanisms and abstract visual patterns, leading to cognitive dissonance.

To mitigate this, we propose a knowledge mining approach guided by HCoT, which emulates the expert reasoning process in SAR interpretation. By progressively integrating multi-source background knowledge and priors, HCoT constructs a structured and controllable cognitive chain, embedded into LLM prompts to enhance text generation quality. The method consists of two key components:

\textit{1) \textbf{HCoT - Knowledge Driven Thinking Chain}:} When interpreting SAR images, human experts typically follow a cognitive path that moves from macro to micro, and from background knowledge to specific goals. Inspired by this process, this study proposes a five-level hierarchical knowledge thinking chain, as shown in \figref{fig5}[a], to guide the step-by-step reasoning of the large model:
\begin{itemize}[leftmargin=1em, itemsep=2pt]
\item \textbf{Earth Cognition Layer}: Guided by the geographic information, we prompt the LLM to activate macro-level understanding of the target region based on its world knowledge. For example, when the area corresponds to Beijing Daxing Airport, the model can infer its role as an international hub, associating it with multi-runway layouts, common aircraft types, and terminal structures. Likewise, industrial zones suggest factories and warehouses, while residential areas imply apartment complexes and supporting infrastructure.
    \item \textbf{Social Prior Information}: Building on geographic context, this layer guides the LLM to integrate economic structure, architectural forms, and transport patterns for functional inference. For example, flight data and urban planning aid in inferring typical aircraft and infrastructure at airports; manufacturing zones imply large factories, while tech parks suggest high-rise offices. Tailored prompts enhance the LLM’s scene-specific cognition across airports, ports, cities, and agricultural regions.
    \item \textbf{SAR Theoretical Knowledge Layer}: This layer guides the LLM to incorporate SAR imaging principles—such as scattering behavior, speckle noise, and top-down inversion—for interpreting unique visual patterns. It provides domain-specific priors like “runways appear as dark strips,” “grasslands exhibit uniform textures,” and “metal objects yield strong scattering,” enabling the LLM to distinguish prominent targets from background clutter. Such physical constraints help associate patterns (e.g., rectangular outlines with buildings, Y-shaped structures with aircraft), thereby grounding semantic understanding in SAR-specific physics.
    \item \textbf{Instance-Level Discrimination Layer}:  Based on macroscopic context, this layer directs the LLM to perform fine-grained recognition of specific targets (e.g., aircraft, buildings, roads). As visual cues alone are often insufficient, SAR spatial resolution is leveraged to support scale-aware reasoning. For instance, fuselage length and wingspan inferred from scattering patterns and resolution can help distinguish a Boeing 737 from an Airbus A350. Likewise, building height and area, estimated via contour shape and shadow extent, assist in differentiating civil from industrial structures. This approach improves the model’s accuracy in target categorization.
    \item \textbf{Knowledge Calibration and Decision-Making Layer}: After completing the first four stages of reasoning, the model has acquired multi-dimensional knowledge. However, due to the inherent complexity of SAR image interpretation, this information requires further cross-validation and integration. The prompt guides the model to calibrate its output based on high-confidence cues—such as geographic context and social priors—ensuring that the generated text aligns with SAR imaging principles while maintaining semantic validity, realism, and logical coherence.
\end{itemize}

\textit{2) \textbf{HCoT - Multi-scale Image Semantic (HCoT-MIS)}:} In Section \ref{sec3.B}, we introduce a spatial-resolution-based tiling strategy, SRC, that unifies semantic granularity across data from diverse SAR sensors. This organization enables complementary semantics at multiple scales: large-scale images capture regional layout and context, medium-scale images highlight target structures and spatial distribution, while small-scale images reveal fine-grained attributes and scattering characteristics.

Building on this, we propose a multi-scale semantic-driven reasoning framework, HCoT-MIS, as illustrated in \figref{fig5}[b]. Its core idea is to construct a coherent cross-scale semantic progression to enable stepwise information flow and fusion, thereby alleviating the semantic fragmentation and logical inconsistencies inherent in traditional single-scale analysis.
HCoT-MIS operates in three reasoning stages, hierarchically organized from large to small spatial scales, and can be formally described as follows:
\begin{equation}
T_L = f\theta(S_L),
\end{equation}
\begin{equation}
T_M = f\theta(S_M, T_L),
\end{equation}
\begin{equation}
T_S = f\theta(S_S, T_M).
\end{equation}

Among them, $S_L$, $S_M$, and $S_S$ represent large-scale, medium-scale, and small-scale SAR images, respectively, while $T_L$, $T_M$, and $T_S$ denote the text descriptions generated at the corresponding scales. The function $f\textit{$\theta$}$ represents the inference function driven by the LLM.

\textbf{In the first stage}, the large-scale image is input into the LLM, where the model’s inherent world knowledge is used to provide a macroscopic understanding and description of the overall geographical background, regional functions, and environmental context of the area covered by the image.

\textbf{In the second stage}, we incorporate the macroscopic background information generated from the large-scale image as prior knowledge, inputting it alongside the medium-scale image into the LLM. This combined information allows the model to leverage existing global context to further refine and accurately describe the structural layout and target distribution within the region when generating the medium-scale image description.

\textbf{In the third stage}, the small-scale image refines the specific attributes and details of the targets (such as type, structure, size, etc.) based on the prior scales.

This mechanism simulates the human spatial cognitive process, moving from global to local, enabling the model to focus on local details when describing the targets, while ensuring consistency and logical coherence with the overall context. Guided by the HCoT reasoning mechanism, more than 1 million structured texts were generated, forming the language modality data of FUSAR-GEOVL-1M. A detailed statistical analysis and example verification are provided in Section \ref{sec4.1}.

\subsection{FUSAR-KLIP Overall Framework}
\begin{figure*}[htbp]
	\centerline{
		\phantomsection
		\includegraphics[width=0.9\textwidth]{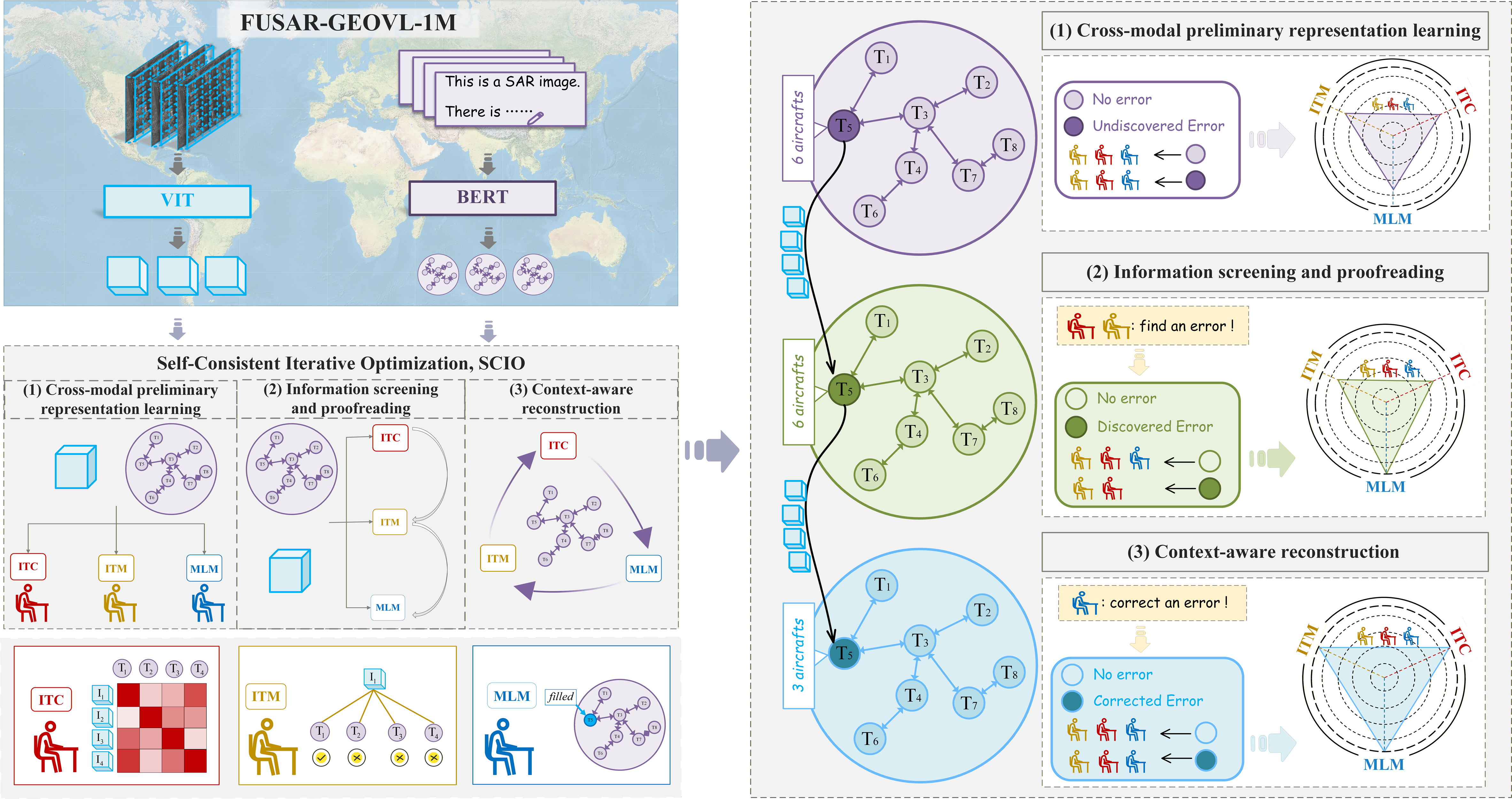}}	
	\caption{The overall framework of FUSAR-KLIP. SAR images and texts from FUSAR-GEOVL-1M are encoded by ViT and BERT, respectively, and optimized through the SCIO module. SCIO consists of three progressive stages: (1) cross-modal preliminary representation learning, (2) information screening and proofreading, and (3) context-aware reconstruction—forming a closed-loop “screen–filter–reconstruct” process that progressively enhances cross-modal alignment and semantic representation.}
	\label{fig7}
\end{figure*}

After introducing geographic information and scale consistency into images and constructing hierarchical cognitive semantics into text, the key challenge is achieving effective alignment and knowledge fusion of cross-modal representations at the model level. To this end, we propose FUSAR-KLIP, a universal visual language foundational model for SAR imagery. This framework, based on the core principles of "multimodal semantic modeling, modality alignment, and self-supervised optimization," aims to eliminate reliance on manual annotation and achieve deep semantic understanding of SAR imagery through cross-modal collaborative learning.

Specifically, FUSAR-KLIP adopts a standardized dual-encoder architecture to separately model visual and textual representations. The visual encoder employs a vision transformer, where SAR images are divided into fixed-size patches and processed through linear projection and positional encoding before being passed to the Transformer. This yields a uniform visual embedding vector: $f_v\in\mathbb{R}^d$. The text encoder is based on BERT, which tokenizes and embeds the input text, producing a corresponding embedding vector: $f_t\in\mathbb{R}^d$ through a multi-layer Transformer. These embeddings are aligned within a shared semantic space. ViT and BERT, both Transformer-based, are widely adopted in multimodal frameworks (e.g., BLIP, ALBEF, OFA\cite{li2021align},\cite{Liu2025Causal}) due to their strong alignment capability, generalization, and downstream task compatibility, making them well-suited as foundational encoders for our model.

To achieve multimodal interaction and semantic alignment, FUSAR-KLIP introduces three collaborative self-supervision tasks to enhance cross-modal modeling capabilities from three levels: modality alignment, semantic matching, and text generation.

\begin{itemize}[leftmargin=1em, itemsep=2pt]
    \item \textbf{Image-Text Contrastive Loss (ITC)}: To establish alignment in the global cross-modal embedding space, we employ the symmetric $infoNCE$ loss to maximize the cosine similarity of matching image-text pairs while minimizing the similarity of non-matching pairs. The specific calculation method is shown in Equation  \ref{for4}, where $s(\cdot)$ represents cosine similarity, and $\tau$ is the temperature coefficient.
    \begin{equation}
    \mathcal{L}_{ITC} = - \log \frac{e^{s(f_v, f_t)/\tau}}{\sum_{t'} e^{s(f_v, f_{t'})/\tau}}.
    \label{for4}
    \end{equation}
    \item \textbf{Image-Text Matching Loss (ITM)}: This loss enhances the model’s ability to understand fine-grained semantics of images and texts. The image-text features are fused through a cross-attention module and input into a matching discriminant head for binary classification. The loss is computed using cross-entropy, as shown in Equation  \ref{for5}, where $y$ represents the matching label of the image-text pair. If the image-text pair is correctly matched, $y$=$1$ ; otherwise, $y$=$0$. $p$ denotes the image-text matching probability predicted by the model.
    \begin{equation}
    \mathcal{L}_{\text{ITM}} = - y \log(p) - (1 - y) \log(1 - p).
    \label{for5}
    \end{equation}
    \item \textbf{Masked Language Modeling Loss (MLM)}: The MLM task further promotes the ability of cross-modal conditional generative modeling, masks the key information tokens in the text, and achieves semantic enhancement reconstruction through a two-stage process. First, the visual-text is jointly encoded, with the visual and text features fused through a cross-attention layer. Then, a Transformer-based decoder is used to reconstruct the masked token in an autoregressive manner. The MLM loss function is defined as:
    \begin{equation}
    \mathcal{L}_{\text{MLM}} = - \sum_{i \in \text{masked}} \log P(w_i | \hat{w}_i, f_v),
    \label{for6}
    \end{equation}
    where $i$ represents the index of all masked positions,$w_i$ represents the ground truth token at the $i-th$ position, $\hat{w}_i$ represents the visible context excluding position $i$, $f_v$ represents the visual modality feature, and $P(w_i | \hat{w}_i, f_v)$ represents the probability of correctly predicting $w_i$ given the context $\hat{w}_i$ and the image feature $f_v$.
\end{itemize}

The three loss functions are aggregated into a unified multi-task objective via weighted summation, encouraging the model to establish cross-modal associations from visual features to high-level semantics. The balancing coefficient $\lambda$ is set to equal weights by default.
\begin{equation}
\mathcal{L}_{\text{total}} = \lambda_1 \mathcal{L}_{\text{ITC}} + \lambda_2 \mathcal{L}_{\text{ITM}} + \lambda_3 \mathcal{L}_{\text{MLM}}.
\label{forLoss}
\end{equation}

\subsection{Self-Consistent Iterative Optimization (SCIO)}

However, the effectiveness of joint optimization still depends heavily on the quality of the input text. Although the HCoT prompt strategy significantly enhances the ability of large language models to understand SAR images and generate usable text, there are still inevitable factual biases. When used directly for cross-modal pre-training, this can introduce noise interference and diminish the effectiveness of feature alignment. To address this, we propose a SCIO module, which establishes a closed-loop "screen–filter–reconstruct" mechanism within a fully self-supervised framework, enabling progressive optimization of language modalities. 

As shown in \figref{fig7}, the SCIO module consists of three progressive stages integrated into training. By emphasizing ITC, ITM, and MLM at different stages, it enables joint optimization that progressively improves text quality and cross-modal alignment. The mechanism is composed of three stages:

\textit{1) \textbf{Cross-Modal Representation Learning:}} In the initial stage of SCIO, we perform multimodal self-supervised pre-training for each SAR image $v_i$ and its corresponding text description $t_i$. The aligned text $t_i$ for each image consists of 8 different sub-texts, covering various information such as scene background, regional function, terrain structure, target type, and layout relationships. We use the visual encoder $f_v(\cdot)$ and the text encoder $f_t(\cdot)$ to extract features from the image and text, respectively:
\begin{equation}
    v_i=f_v(Image_i),
    \label{for7}
\end{equation}
\begin{equation}
    t_i=f_t(Text_i).
    \label{for8}
\end{equation}

In this stage, the model is jointly optimized based on Equation  \ref{forLoss}. However, since the generated text inevitably contains noise, the training at this stage typically only captures limited cross-modal associations. Noisy text not only interferes with the learning quality of ITC and ITM but also reduces the language modeling accuracy of MLM. To address this, in the subsequent stage, we introduce a sample screening mechanism based on ITC and ITM responses to eliminate low-confidence texts, improving the reliability of the input corpus and providing MLM with purer and more reliable training data.

\textit{2) \textbf{Text Screening and Proofreading:}} This stage introduces a segment-level noise removal mechanism further to improve the representation quality of the language modality. We perform screening for each image's 8 sub-text segments \{$p_1$,...,$p_8$\} based on the ITC and ITM tasks.

For each candidate text segment $p_j$, we construct a new version of the text $t_i^{(-j)}$ by removing the segment, and calculate the corresponding image-text ITC contrast loss difference:
\begin{equation}
    \Delta L^{(j)}_{\text{ITC}} = L_{\text{ITC}}(v_i, t_i^{(-j)}) - L_{\text{ITC}}(v_i, t_i).
\end{equation}

If $\Delta L^{(j)}_{ITC} < 0$, the removal of the segment improves image-text alignment, identifying it as potential noise. These candidate noise segments are further evaluated by computing the change in image-text matching scores via the ITM module:

\begin{equation}
    \Delta L^{(j)}_{\text{ITM}} = s_i^{(-j)}-s_i,
\end{equation}
Where $s_i^{(-j)}$ is the image-text matching score after removing the fragment, and $s_i$ is the score of the original text. If $\Delta L^{(j)}_{ITM}>0$, the fragment is further confirmed as noise, added to the noise pool, and excluded from the subsequent MLM training.

At this stage, ITC and ITM losses are still computed using the complete original text to maintain stability and consistency in the optimization objective. However, noisy segments are excluded from the MLM prediction targets in the subsequent stage, thereby reducing semantic interference and enhancing the accuracy and robustness of language modeling.

	\begin{algorithm}[t]
		
		\footnotesize
		\caption{Self-Consistent Iterative Optimization}
		\label{alg:scio}
		
		\KwIn{$I$: SAR image, $T = \{t_1, t_2, \dots, t_8\}$: Text, $\theta$: Model parameters}
		\KwOut{$T_{\text{final}}$: Optimized textual description aligned with SAR image}
		
		\textbf{Initialize} model parameters $\theta$\\
		
		\For{$\text{iteration} = 1$ \KwTo max\_iter}{
			\textbf{\textcolor{blue}{Phase 1: Pretraining and Alignment}}
			\[
			\mathcal{L} = \mathcal{L}_{\text{ITC}} + \mathcal{L}_{\text{ITM}} + \mathcal{L}_{\text{MLM}}
			\]

			\textbf{\textcolor{blue}{Phase 2: Text Filtering and Refinement}}\\
			\For{each sub-sentence $t_i \in T$}{
				\[
				\Delta \mathcal{L}_{\text{ITC}}(t_i) = 
                \mathcal{L}_{\text{ITC}}(T-t_i, I) - 
                \mathcal{L}_{\text{ITC}}(T, I)
				\]
				\If{$\Delta \mathcal{L}_{\text{ITC}}(t_i) < 0$}{
					Remove $t_i$ from $T$ $\rightarrow$ $T'$
				}
			}
			Refine text $T'$ using ITM:  
			\[
			\mathcal{L}_{\text{ITM}}(T', I) < \mathcal{L}_{\text{ITM}}(T, I) \quad \text{Accept $T'$ if true}
			\]
			
			\textbf{\textcolor{blue}{Phase 3: Contextual Reconstruction with MLM}}\\
			Mask tokens in $T'$ and predict missing tokens to generate the reconstructed text $T_{\text{filled}}$:  
			\[
			T_{\text{filled}} = \text{MLM-decoder}(T_{\text{masked}}, I)
			\]
			
			Evaluate using ITC and ITM:  
			\[
			\mathcal{L}_{\text{ITC}}(T_{\text{filled}}, I) < \mathcal{L}_{\text{ITC}}(T', I)  
			\]
			\[
			\mathcal{L}_{\text{ITM}}(T_{\text{filled}}, I) < \mathcal{L}_{\text{ITM}}(T', I)
			\]
			\If{both conditions hold}{
				$T_{\text{final}} = T_{\text{filled}}$
			}
			\Else{
				$T_{\text{final}} = T'$
			}
			
			\textbf{Feedback Loop: Backpropagate optimized text}\\

		}
		
		\KwOut{$T_{\text{final}}$}
	\end{algorithm}

\textit{\textbf{(3) Context-Aware Reconstruction:}} After the first two stages of screening and optimization, the text decoder in the MLM task is trained on a cleaner corpus, improving its capacity for image-grounded understanding and generation. Leveraging fused image-text features, the decoder reconstructs masked, noisy fragments using contextual and visual cues, enhancing overall text quality. Specifically, each noisy fragment $p_j$ is replaced with a $[mask]$ token during reconstruction:
\begin{equation}
    t_j^{mask}=\{p_1,...,p_{j-1},[mask],p_{j+1},...,p_n\}.
\end{equation}

Then, $t_j^{mask}$ and the image feature $v_i$ are input into the text and visual encoders together to reconstruct the missing fragment based on the cross-modal context. $\hat{t_i}$ represents the text after the missing fragment has been generated by MLM:
\begin{equation}
    \hat{t_i}=MLM(t_j^{mask},v_i).
\end{equation}

The text generated by the MLM is further evaluated to determine whether it improves upon the original text:
\begin{equation}
    \Delta L^{(\hat{p_j})}_{\text{ITC}} = L_{\text{ITC}}(v_i, \hat{t_i}) - L_{\text{ITC}}(v_i, t_i),
\end{equation}
\begin{equation}
    \Delta L^{(\hat{p_j})}_{\text{ITM}} = \hat{s_i}-s_i.
\end{equation}

If $\Delta L^{(\hat{p_j})}_{ITC}<0$ and $\Delta s^{(\hat{p_j})}_{\text{ITM}}>0$ are satisfied, it indicates that the reconstructed fragment outperforms the original fragment in both image-text alignment and semantic matching. In this case, we replace the original fragment $p_j$ in the text with the reconstructed fragment $\hat{p_j}$ , and train the model based on this.

Overall, the SCIO module employs a collaborative mechanism of progressive optimization and closed-loop feedback: the first two stages use ITC and ITM to filter out high-quality text subsets, thereby enhancing the purity of the MLM training data. Meanwhile, the MLM reconstruction module optimizes the image-text alignment task to achieve high-quality semantic complementarity between text and image representations. Algorithm \ref{alg:scio} outlines the overall process of SCIO.

\begin{figure*}[htbp]
	\centerline{
		\phantomsection
		\includegraphics[width=0.9\textwidth]{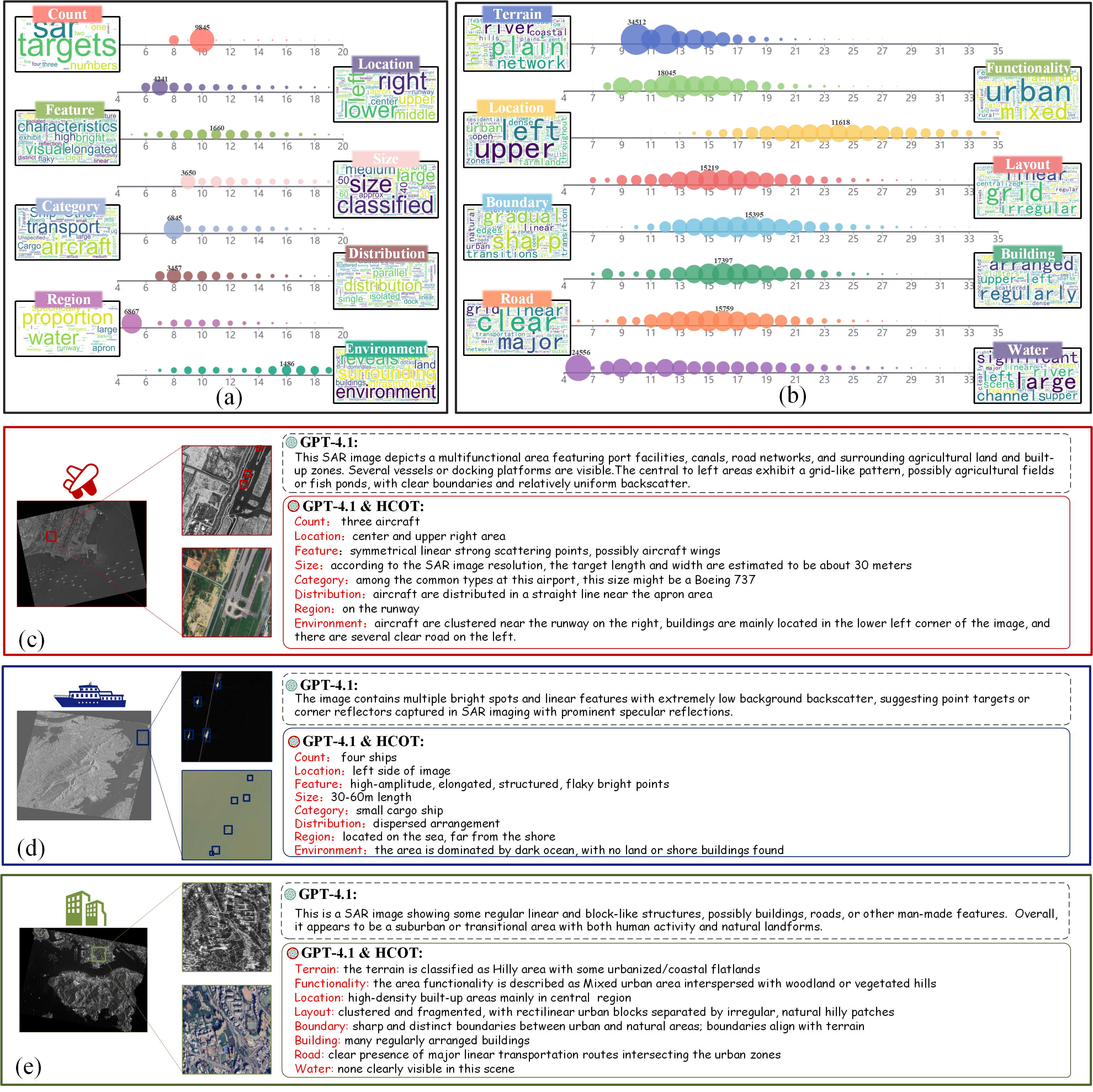}}	
	\caption{Text data of FUSAR-GEOVL-1M. (a)–(b): word clouds and length statistics of object and landform task texts; (c)–(e): Influence of HCoT on text quality.}
	\label{fig8}
\end{figure*}

\section{Experiments and analysis}\label{sec4}
\subsection{Implementation Details}\label{sec4.1}
FUSAR-KLIP relies on large-scale cross-modal pre-training, which requires balancing computational efficiency with model capacity while ensuring reproducibility. Accordingly, we implemented the complete training and evaluation pipeline in PyTorch and conducted all experiments on a computing node equipped with eight NVIDIA RTX 3090 GPUs. The visual encoder is initialized from a ViT model pretrained on ImageNet, while the text encoder uses BERT weights. Considering computational efficiency and model performance, the input image size is fixed at 224×224 with a batch size of 32 during pretraining. A queue of 20,000 image-text features is maintained to support large-scale contrastive learning. The alignment loss weight $\alpha$ is set to 0.4 to strengthen cross-modal semantic association.

 \begin{table}[!htbp]
\small
\caption{Current Advanced Remote Sensing Multimodal Models}
\centering
\setlength{\tabcolsep}{5pt}
\renewcommand{\arraystretch}{1.4}
\resizebox{0.48\textwidth}{!}{
\begin{tabular}{l l c c}
\hline

\textbf{Model} & \textbf{Vision Backbone} & \textbf{Publication} & \textbf{Image-Text Pairs} \\ \hline
RemoteClip\cite{10504785}  & ViT-Base\textbackslash Large  & TGRS-2024      &165,754    \\ 
GeoRSClip\cite{10679571}    & ViT-Base          & TGRS-2024      & 5,070,186  \\ 
BAN\cite{10438490}          & ViT-Base\textbackslash Large   & TGRS-2024      & 23,822    \\ 
ChangeClip\cite{dong2024changeclip}   & ViT-Base          & ISPRS-2024     & 59,246     \\ 
Prithvi\cite{10641903}      & ViT-Base          & IGARSS-2024    & 593,082   \\ 
Geochat\cite{kuckreja2024geochat}      & ViT-Large          & CVPR-2024      & 141,246    \\ 
VHM\cite{pang2025vhm}          & ViT-Large          & AAAI-2025      & 1,390,405   \\ 
SkyScript\cite{wang2024skyscript}    & ViT-Base\textbackslash Large   & AAAI-2024      & 2,600,000   \\ 
BITA\cite{10415446}         & ViT-Large          & TGRS-2024      & 44,521     \\ 
SARCLIP\cite{jiang2026sarclip}         & ViT-Base\textbackslash Large          & ISPRS-2026      & 400,000     \\ 
CLIP*\cite{radford2021learning}         & ViT-Base          & ICML-2021      & 400 M       \\ 
BLIP*\cite{li2022blip}         & ViT-Base          & ICML-2022      & 100 M       \\ \hline
\multicolumn{4}{l}{\scriptsize\textit{Note:} models marked with * are trained on natural image data.} \\
\end{tabular}
}
\label{tab2}
\end{table}

The model is optimized using AdamW with a weight decay of 0.05. The learning rate is initialized at 3e-4, linearly warmed up from 1e-6 over 3000 steps, and then decays to a minimum of 1e-6 following a rate of 0.9. Training is conducted over 36 epochs, with each stage of SCIO trained for 12 epochs.

For downstream task evaluation, we systematically fine-tuned the proposed foundation model across 11 representative remote sensing tasks in the vision and language categories, comprehensively examining its generalization and practical performance. In this process, we constructed a SAR multimodal evaluation benchmark and compared it with several leading open-source remote sensing multimodal models listed in Table~\ref{tab2}, rigorously evaluating them under a unified setup. Due to the scarcity of high-quality SAR image-text pairs, these models are mostly pre-trained on natural or optical remote sensing imagery. This benchmark not only provides a systematic validation framework for this research but also establishes a reusable evaluation baseline for subsequent related work.

\subsection{Analysis of FUSAR-GEOVL-1M Text}

After constructing the FUSAR-GEOVL-1M dataset, we further conducted statistical and quality analysis of its language modalities. Unlike existing multimodal remote sensing datasets that rely on manual or templated labeling, the data text in this study was automatically generated by a large language model, necessitating a systematic assessment of its semantic richness, task relevance, and diversity. To ensure the accuracy and interpretability of the generated text, we constructed semantic descriptions based on GPT-4.1. This model performed best in SAR scene understanding, terminology usage, and semantic coherence in comparisons with mainstream language models like Gemini and Grok, and was therefore selected as the primary generation engine.

During dataset construction, we designed task-oriented prompts based on the HCoT framework introduced in Section \ref{sec3.c}, aiming to enhance the reasoning capability and fine-grained perception of large language models in SAR image interpretation. Instruction sets were developed for three representative remote sensing tasks: ground target recognition, marine target recognition, and terrain understanding.
For target-related tasks, prompts guide the model to analyze attributes such as location, quantity, category, and structural features. In terrain-oriented tasks, the focus is on interpreting elements such as buildings, water bodies, and road networks. Each SAR image is paired with eight complementary textual descriptions spanning multiple semantic levels and knowledge dimensions. Key information is extracted via regular expression matching, resulting in a total of one million structured descriptions.

\figref{fig8}[a,b] present the word frequency distributions and length statistics for target and terrain descriptions, respectively. Terrain-related prompts yield more diverse and open-form responses, reflected in denser textual distributions. \figref{fig8}[c,d,e] showcase representative samples from FUSAR-GEOVL-1M, corresponding to ground targets, marine targets, and terrain scenarios. The integration of the HCoT prompting strategy leads to a notable improvement in GPT-4.1’s reasoning performance and semantic expressiveness.

\begin{figure}[!htbp]
	\centerline{
		\phantomsection
		\includegraphics[width=0.4\textwidth]{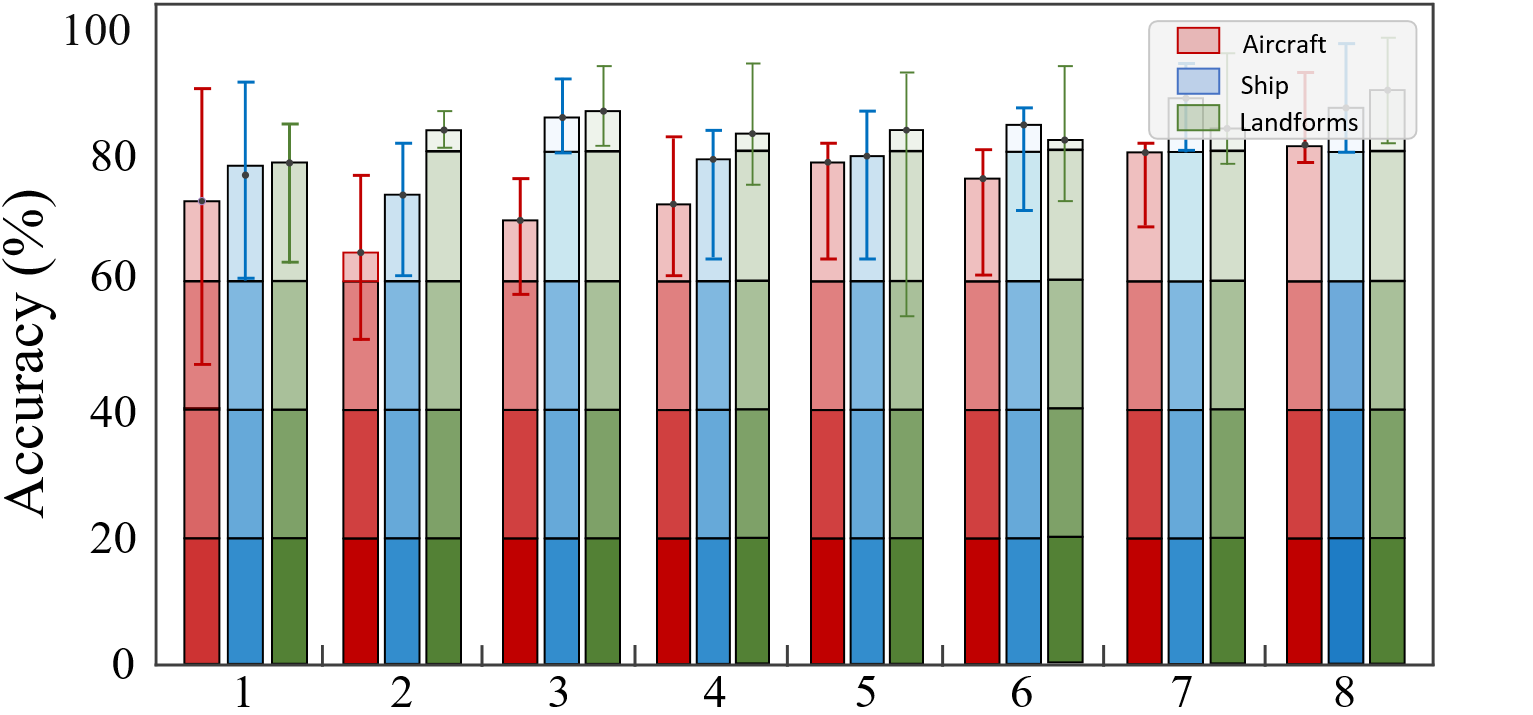}}	
	\caption{Accuracy evaluation of eight dimensions of information in FUSAR-GEOVL-1M text data.}
	\label{fig_txt_acc}
\end{figure}

To assess the textual accuracy of FUSAR-GEOVL-1M, we conducted an expert evaluation on a randomly sampled 2\% subset of the dataset. Three experts independently rated eight semantic attributes per image, assigning information accuracy scores to quantify factual consistency. As shown in \figref{fig_txt_acc}, the overall accuracy approaches 80\%. Marine target descriptions achieve higher accuracy than ground targets, benefiting from the relatively uniform sea background. Environmental descriptions also perform well, aided by geospatial cues and lower semantic complexity. While fine-grained target attributes occasionally contain errors, the generated texts are generally accurate and coherent, offering strong support for high-level scene understanding and multimodal pre-training.

\begin{figure}[!tbp]
	\centerline{
		\phantomsection
        \hspace{-0.4cm}
		\includegraphics[width=0.35\textwidth]{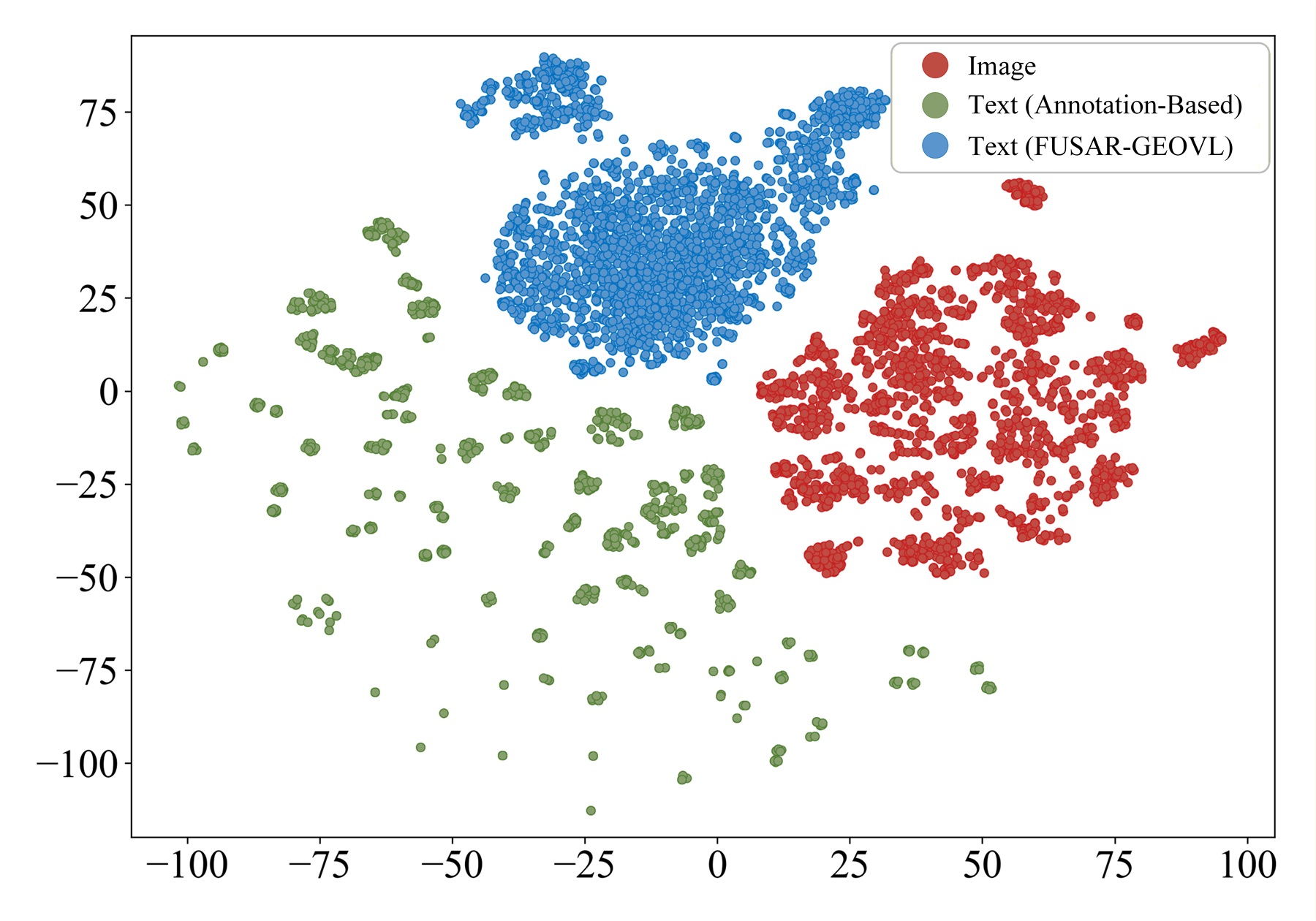}}	
	\caption{TSNE feature distribution of images and two types of text. The text in FUSAR-GEOVL is closer to the image feature distribution, while the distribution of annotation-based text is sparse.}
	\label{fig9a}
\end{figure}

\begin{figure}[!tbp]
	\centerline{
		\phantomsection
		\includegraphics[width=0.4\textwidth]{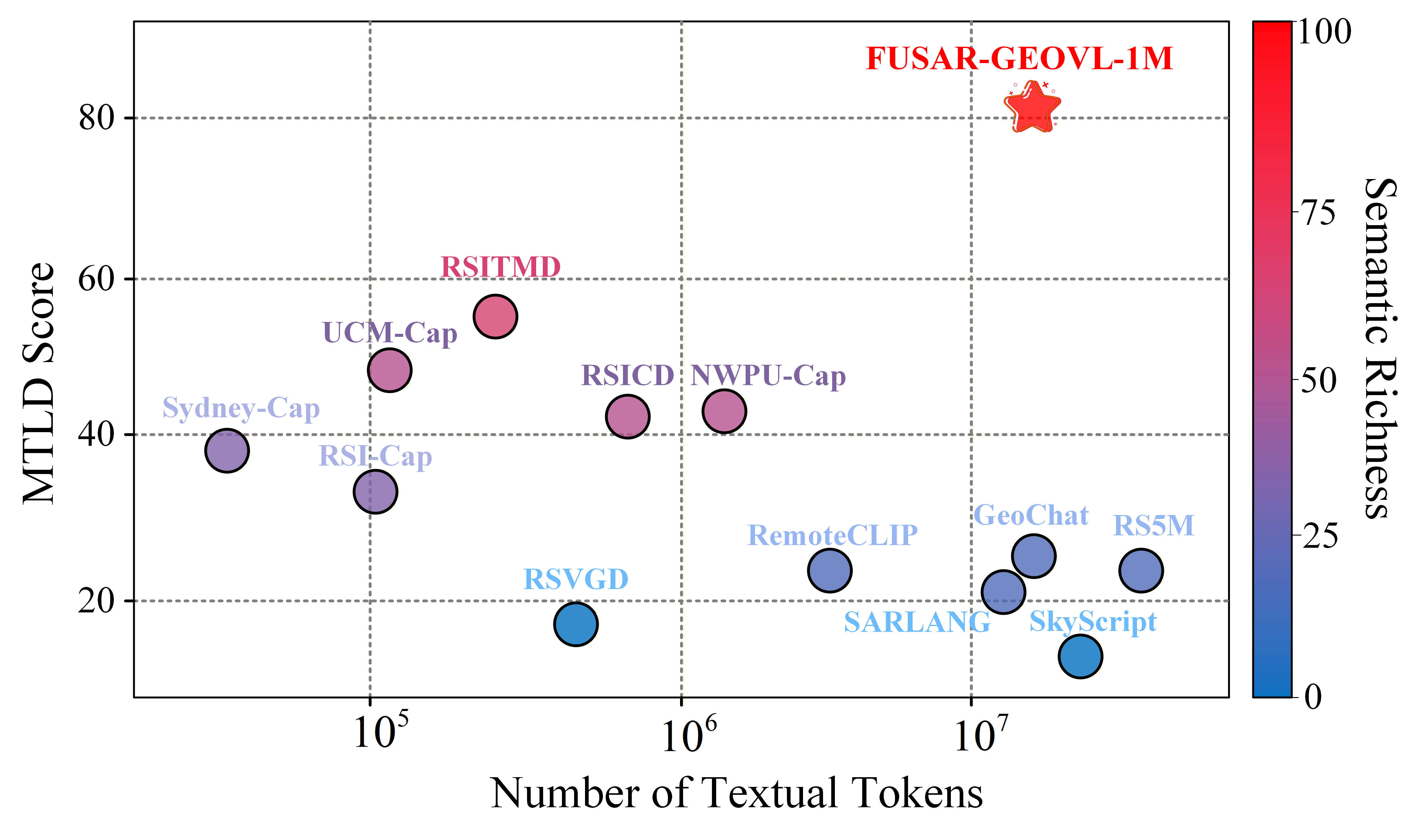}}	
	\caption{The number of text tokens and information richness of public remote sensing multimodal datasets. }
	\label{fig9b}
\end{figure}

To further compare semantic expressiveness, we selected a subset of FUSAR-GEOVL-1M with manual target annotations and generated templated descriptions using the method in RemoteCLIP~\cite{10504785}. As shown in \figref{fig9a}, t-SNE visualization reveals that FUSAR-GEOVL-1M texts exhibit tighter clustering aligned with image features, whereas templated texts are more dispersed, indicating semantic sparsity. Additionally, as illustrated in \figref{fig9b}, we compared and analyzed multiple remote sensing multimodal datasets using two metrics: the text semantic richness index (MTLD\cite{mccarthy2005assessment}) and token count. Overall, FUSAR-GEOVL-1M significantly outperforms existing mainstream datasets in both semantic richness and token count.

\subsection{Visual Task Benchmarks}

Visual representation capabilities are the cornerstone of cross-modal modeling. For SAR imagery, texture and geometric features under complex electromagnetic scattering mechanisms often directly determine the effectiveness of target recognition and scene parsing. Therefore, before moving on to cross-modal tasks, we first systematically evaluated the model's performance in a single visual modality to verify its ability to support basic perception tasks. Specifically, we selected three typical tasks for evaluation: classification, detection, and segmentation\cite{10509806},\cite{8917599},\cite{10188509}. These tasks cover different levels of requirements, from the object level to the scene level. The three tasks are based on ViT as the visual encoder to build corresponding models, and targeted optimization adjustments are made.

All models use the visual encoder weights in the multimodal foundation model as initialization parameters for subsequent SFT. In addition, three single-modal VIT foundation models are also used for experimental comparison. Among them, MAE\cite{he2022masked} and MOCO v3\cite{Chen_2021_ICCV} are pre-trained based on the dataset proposed in this paper, while SAR-JEPA\cite{LI2024326} is the foundation model proposed by Li et al. for SAR image interpretation. Table \ref{tab3} summarizes the key model configurations and training hyperparameters of each subtask. The training and test data are divided in a ratio of 4:1 by default.

\begin{table}[htbp]
	\setlength{\tabcolsep}{3pt}

	\centering
	\caption{Experimental Setup for Vision Tasks}
	\resizebox{0.48\textwidth}{!}{  
		\renewcommand{\arraystretch}{1.5}
		\begin{tabular}{ l c c c c c c }
			\hline
			\textbf{Task} & \textbf{Model}  & \textbf{Optimizer} & \textbf{LR} & \textbf{Epoch} & \textbf{Scheduling} \\ \hline
			\textbf{Classification} & ViT-Cls\cite{dosovitskiy2021an}  & AdamW & 0.003 & 100 & Warmup,CosineDecay \\ 
			\textbf{Detection} & ViTDET\cite{li2022exploring}  & AdamW & 0.0001 & 36 & Warmup,MultiStep \\ 
			\textbf{Segmentation} & Segmenter\cite{strudel2021segmenter} & AdamW & 0.001 & 100 & PolyLR \\ \hline
	\end{tabular}}
	\label{tab3}
\end{table}

\begin{table*}[!b]
	\centering
	\small
	\setlength{\tabcolsep}{7pt}
	\caption{Benchmarks for vision tasks: target classification, target detection, and segmentation. The top half of the table shows the model using vit-base as the backbone. The bottom half shows the model using vit-large as the backbone. Suboptimal results are \underline{underlined.}}
	\resizebox{1\textwidth}{!}{
		\renewcommand{\arraystretch}{1.4}
		\begin{tabular}{l c|c c|c c|c c|c|c|c|c c|c c}
			\hline
			\multicolumn{2}{c|}{\textbf{Task}} & \multicolumn{6}{c|}{\textbf{Target Classification }}&\multicolumn{3}{c|}{\textbf{Target  Detection}}&\multicolumn{4}{c}{\textbf{Segmentation}}\\\hline
			\multirow{2}{*}{\textbf{Pretrain Model}} & \multirow{2}{*}{\textbf{Backbone}} & \multicolumn{2}{c|}{\textbf{FUSAR-AIR}} & \multicolumn{2}{c|}{\textbf{FUSAR-SHIP}} & \multicolumn{2}{c|}{\textbf{SAR-ACD}} & \multirow{2}{*}{\makecell{\textbf{FU-AS}\\\textbf{}}}&\multirow{2}{*}{\makecell{\textbf{FU-SS}\\\textbf{}}}&\multirow{2}{*}{\makecell{\textbf{AIR-F}\\\textbf{}}}
			&\multicolumn{2}{c|}{\textbf{FUSAR-MAP}} & \multicolumn{2}{c}{\makecell{\textbf{PoSAR-Seg}\\}}\\ 
			&                           & \textbf{Top1} & \textbf{Top3} & \textbf{Top1} & \textbf{Top3} & \textbf{Top1} & \textbf{Top3} & \textbf{mAP} &  \textbf{mAP}& \textbf{mAP} & \textbf{OA} & \textbf{mIoU} & \textbf{OA} & \textbf{mIoU} \\ \hline
			RemoteClip                       & ViT-Base                 & 66.76   & 87.68   & 64.63   & 95.27   & 44.57   & 82.73     & 65.36                    & 81.41                    & 51.31    & 78.38   & 38.09 & 68.12 & 44.18                \\ 
			GeoRSClip                        & ViT-Base                 & 59.64   & 83.97   & 65.05   & 92.86   & 44.57   & 83.88     & 67.94                    & 81.51                    & 53.83    & 78.03   & 37.64 & 64.29 & 44.43                \\  
			BAN                               & ViT-Base                 & 65.28   & 85.90   & 65.79   & 95.27   & 37.00   & 76.64     &  67.97                   & 80.47                    & 51.94    & 78.56   & 38.15 & 64.74 & 44.53                \\  
			ChangeClip                        & ViT-Base                 & 56.97   & 83.08   & 65.79   & 95.06   & 33.22   & 71.38     & 66.98                    & 81.77                    & 50.96    & 78.28   & 38.05 & 66.63 & 43.31                \\  
			Prithvi                            & ViT-Base                 & 67.50   & 89.31   & 64.63   & 93.28   & 54.93   & 91.61     & 57.66                    & 78.38                    & 35.87    & 78.59   & 38.05 & 63.28 & 36.03                \\ 
			SkyScript                         & ViT-Base                 & 62.31   & 85.45   & 65.58   &  \underline{95.48}   & 54.11   & 88.48     & 66.91                    & 80.86                    & 55.99    & 78.58   & 38.19 & 64.89 & 41.79                \\ 
			CLIP                              & ViT-Base                 & 62.61   & 85.60   & 64.21   & 94.12   & 38.65   & 83.05    & 66.82                    & 81.57                    & 48.43       & 78.39   & 38.41 & 74.35 &  \underline{50.48}              \\ 
			BLIP                              & ViT-Base                 & 64.98   & 86.05   & 65.58   & 94.27   & 56.25   & 92.59    & 67.86                    &  \underline{83.18}                    & 52.84      & 79.32   & 40.69 & 51.88 & 20.05               \\  
			SAR-JEPA                          & ViT-Base                 & 74.33   &  \underline{90.94}   & 63.69   & 94.54   & 69.40   &  \underline{95.72}    & 54.72                    & 78.65                    & 48.02      & 78.63   & 38.15 & 64.61 & 35.98                \\ 
			MAE                               & ViT-Base                 & 66.32   & 87.83   &  \underline{67.89}   & 94.22   & 63.81   & 92.43     & 67.05                    & 82.31                    & 58.01    &  \underline{79.65}   &  \underline{41.23} & 72.46 & 46.27                 \\ 
			MoCo v3                           & ViT-Base                 & \underline{75.51}   & 90.65   & 64.21   & 93.17   &  \underline{71.87}   &  \underline{95.72}    & 67.84                    & 82.20                    &  \underline{58.90}     & 79.53   & 40.71 &  \underline{74.55} & 50.37                \\ 
            SARCLIP& ViT-Base& 66.32& 86.35&66.94 &95.35& 41.44& 80.75& \underline{68.20}&80.91&53.92 &78.38  &37.83  &64.63 &38.14 \\
			\hline
			\rowcolor{lightgray}\textbf{FUSAR-KLIP}               & \textbf{ViT-Base}       & \textbf{81.90}   & \textbf{93.32}   & \textbf{69.15}   & \textbf{96.54}  & \textbf{91.11}  & \textbf{99.84}  & \textbf{74.36}           & \textbf{85.87}           & \textbf{73.04}& \textbf{81.37} & \textbf{43.01} & \textbf{76.75} & \textbf{51.75} \\ \hline 
			RemoteClip                       & ViT-Large                & 65.28   &  \underline{87.38}   &  \underline{61.69}   & 92.54   & 45.55   & 87.00     & 58.87                    & 77.01                     & 51.56   & 61.93   & 19.87 & 62.65 & 30.82                 \\ 
			BAN                               & ViT-Large                & \underline{65.87}   &  87.37   & 55.71   & 90.97   &  \underline{51.31}   & 89.14     & 59.25                    & 77.06                     & 49.82   & \underline{78.48}   &  \underline{38.13} & 61.64 & 33.48                 \\  
			GeoChat                           & ViT-Large                & 62.46   & 86.35   & 60.70   & 91.81   & 43.25   & 80.92    & 59.01                     & 77.13                    & 41.17     & 76.15   & 38.11 & 44.80 & 19.70                \\ 
			VHM                               & ViT-Large                & 60.68   & 84.12   &  \underline{61.69}   & 92.75   & 30.92   & 69.73     & 57.83                    &  78.08                    & 44.19   & 67.18   & 27.34 & 47.66 & 17.56                 \\ 
			SkyScript                         & ViT-Large                & 65.13   & 86.50   & 65.58   &  \underline{94.59}   & 47.53   &  \underline{89.47}    & 58.72                    & 77.87                    & 45.93     & 72.60   & 31.76 &  \underline{66.02} & 43.20                \\ 
			BITA                              & ViT-Large                & 62.61   & 86.94   & 57.08   & 89.92   & 37.00   & 80.26    &  \underline{62.35}                    & 77.25                    & 45.80     & 58.78   & 16.34 & 65.40 &  \underline{44.58}                \\  
            SARCLIP& ViT-Large& 60.68 & 85.46 & 51.52 &89.71 & 48.35 &87.99  &62.10  & \underline{78.63}& \underline{60.28}&63.84 & 17.55 &63.66 &35.39  \\
			\hline
			\rowcolor{lightgray}\textbf{FUSAR-KLIP}               & \textbf{ViT-Large}      & \textbf{75.96}   & \textbf{90.35}   & \textbf{66.84}   & \textbf{95.14} & \textbf{77.46}  & \textbf{98.02}  & \textbf{68.14}                    & \textbf{82.54}                    & \textbf{68.97}              & \textbf{79.13} & \textbf{38.41} & \textbf{70.62} & \textbf{46.29}      \\ \hline
		\end{tabular}
	}
	\label{tab4}
\end{table*}

\textbf{\textit{1) Target Classification:}} The classifier is implemented using the PyTorch framework and evaluated on three representative SAR datasets: FUSAR-SHIP~\cite{hou2020fusar}, FUSAR-AIR~\cite{Qian2024Thesis}, and SAR-ACD~\cite{9754573}. FUSAR-SHIP and FUSAR-AIR are curated subsets of FUSAR-GEOVL-1M, focusing on ship and aircraft target recognition, respectively. FUSAR-SHIP, developed by the Key Laboratory of Electromagnetic Wave Information Science at Fudan University, includes 5,242 images spanning 15 ship categories and 98 subcategories. FUSAR-AIR comprises diverse aircraft types such as transport, refueling, and civil aircraft. SAR-ACD, a public third-party dataset with higher scene complexity and 1-meter resolution, contains 3,032 images across six aircraft types, with approximately 500 samples per class.

Table~\ref{tab4} presents the classification results on the three SAR datasets, evaluated using Top-1 and Top-3 accuracy. Top-1 accuracy reflects the proportion of samples where the model’s top prediction matches the ground truth, while Top-3 accuracy assesses whether the correct label appears within the top three predicted categories, providing a more comprehensive view of model performance across varying scenarios.

As shown in Table~\ref{tab4}, the proposed FUSAR-KLIP foundation  model consistently outperforms all baselines in aircraft and ship classification tasks, demonstrating superior recognition capability. On the SAR-ACD dataset, it achieves a Top-3 accuracy of 99.84\%, highlighting its strength in fine-grained category discrimination. While SAR-JEPA, specifically designed for SAR classification, also delivers strong performance. MoCo v3 and MAE, both pre-trained on SAR data, outperform most multimodal models, highlighting the benefits of modality-specific pretraining. Notably, ViT-Large underperforms due to limited SAR fine-tuning data, whereas FUSAR-KLIP maintains the best performance at equal model scale, reflecting its robustness and generalization ability.

\textbf{\textit{2) Target Detection:}} This study developed an object detection model based on the typical ViT detection framework, ViTDet. By combining the ViT backbone with the feature pyramid structure, ViTDet enhances the detection of multi-scale targets while maintaining global modeling capabilities, achieving excellent performance across various visual detection tasks. We evaluated the model's performance using three representative SAR image datasets, under the AP@50 metric. FUSAR-SHIP-Sense (FU-SS) consists of FUSAR-SHIP target slices expanded into complete scene images, totaling 3,838 images. FUSAR-AIR-Sense (FU-AS) is composed of FUSAR-AIR target slices expanded into complete scene images, totaling 2,491 images. SAR-AIRcraft-Few (AIR-F)\cite{Wang2023SARAIRcraft} is an open-source aircraft detection dataset built from Gaofen-3  images, containing 4,368 images, 16,463 targets, and 7 aircraft categories. This experiment used only 20\% of the training data to assess the generalization ability of the pre-trained model in non-homologous and few-sample scenarios.

\begin{figure}[!tbp]
	\centerline{
		\phantomsection
		\includegraphics[width=0.45\textwidth]{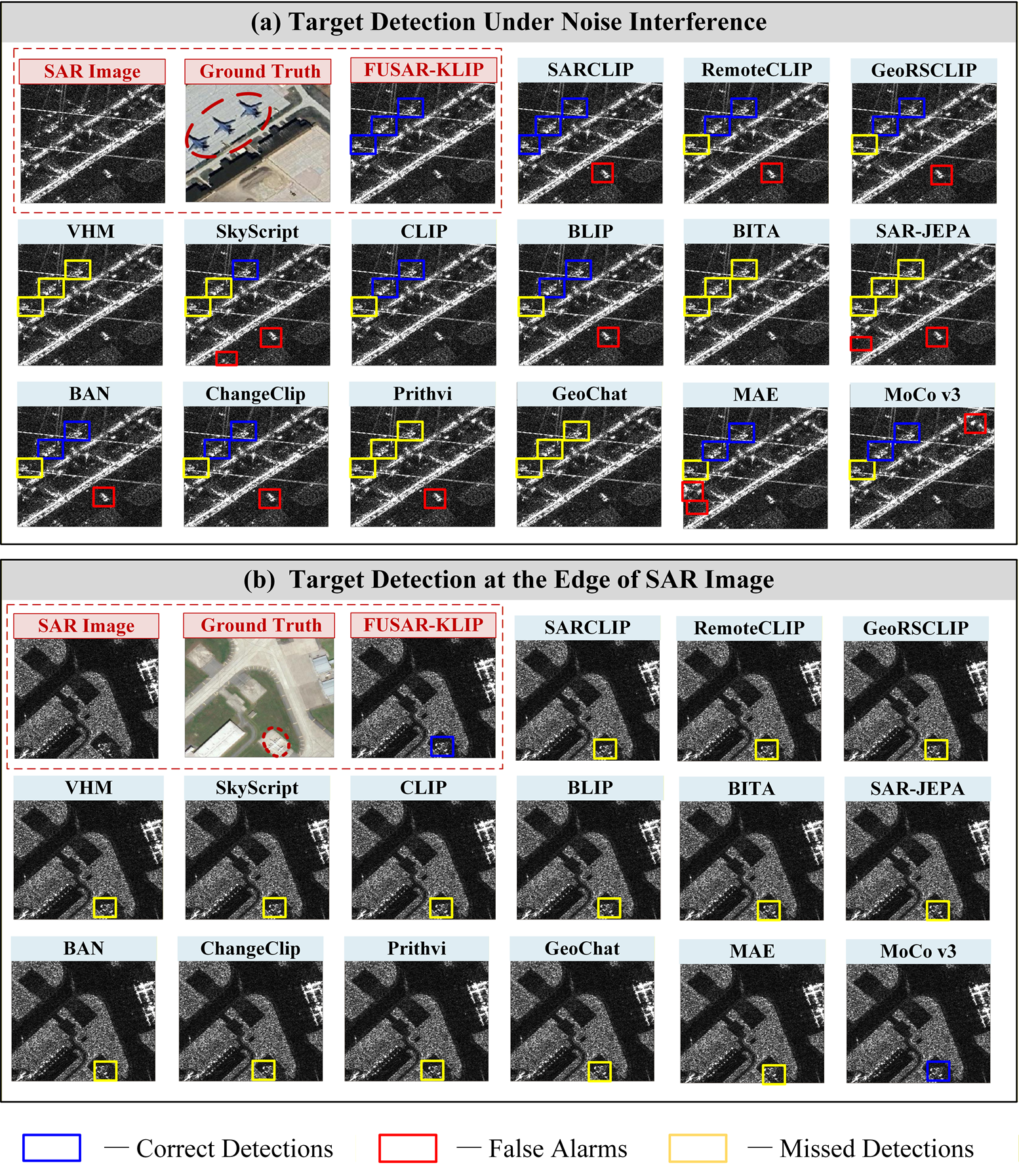}}	
	\caption{Visualization results of the target detection task. \textcolor{blue} {Blue} represents a correct detection. \textcolor{red}{Red}  represents a false alarm. \textcolor{DarkYellow}{Yellow} represents a missed detection.} FUSAR-KLIP has the best performance.
	\label{fig11}
\end{figure}

From the target detection results presented in Table \ref{tab4}, the FUSAR-KLIP model demonstrates significantly superior performance in the three tasks of aircraft detection, ship detection, and few-shot detection, exhibiting excellent target perception and discrimination capabilities. In contrast, GeoRSClip, BAN, BLIP, and MoCo v3 ranked second in some tasks but showed imbalanced performance across other tasks, indicating deficiencies in task generalization and robustness.

\textbf{\textit{3) Segmentation:}} This study constructed a SAR image semantic segmentation model based on PyTorch, employing ViT as the visual encoder and the Mask Transformer from Segmenter as the decoder to enhance semantic modeling. The model was evaluated using overall accuracy (OA) and mean Intersection over Union (mIoU) on two representative datasets. FUSAR-MAP\cite{shi2021object} focuses on fine-grained segmentation of urban scenes—including buildings, roads, vegetation, and background—with 600 high-resolution SAR images (1024×1024). AIR-PolarSAR-Seg (PoSAR-Seg)\cite{wang2022air} targets six land cover categories across 2,000 image slices (512×512), enabling evaluation of general semantic segmentation performance.

The experimental results are presented in Table \ref{tab4}. FUSAR-KLIP achieved the highest $OA$ and $mIoU$ on both segmentation datasets, significantly outperforming existing multimodal and unimodal methods. Among the multimodal models, performance varied greatly due to differences in modeling capabilities of remote sensing domain knowledge. Models such as RemoteClip, GeoRSClip, CLIP, and SkyScript performed relatively well. The unimodal model SAR-JEPA performed well on FUSAR-MAP but showed a significant decline in performance on the more complex AIR-PolarSAR-Seg.

\figref{radar}[a] shows the performance distribution of the models across three visual tasks. In general, FUSAR-KLIP demonstrates stronger generalization ability in typical visual task scenarios by jointly modeling image and text modalities. \figref{fig11} visualizes the detection results for qualitative analysis. For the same model with different ViT sizes, we show better results. \figref{fig11}[a] demonstrates the detection capability of SAR under noise interference. GeoChat, VHM, BITA, and SAR-JEPA all missed targets, while other methods exhibited noticeable false alarms. Only our method successfully detected all targets. \figref{fig11}[b] highlights the detection capability for small targets at the edges. Only our method and MoCo v3 correctly detected these targets. Combining \figref{radar}[a] and \figref{fig11}, FUSAR-KLIP exhibits superior performance in visual tasks.

\begin{figure*}[!tbp]
	\centerline{
		\phantomsection
		\includegraphics[width=0.73\textwidth]{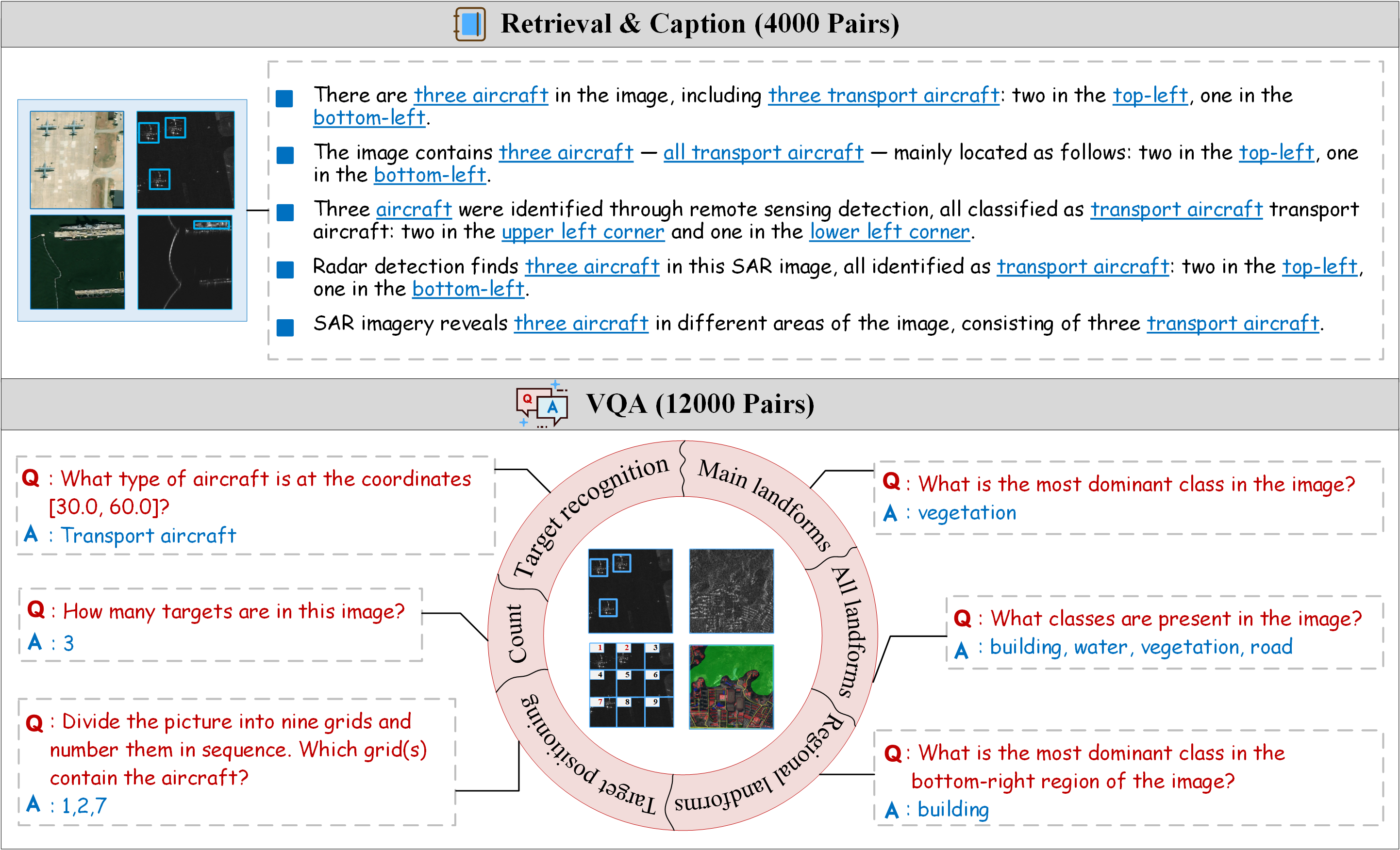}}	
	\caption{Supervised fine-tuning dataset construction for visual language tasks. Captioning and retrieval use the same format of data, and VQA designs 6 tasks for objects and landscapes.}
	\label{fig12}
\end{figure*}

\subsection{Visual-Language Tasks Benchmarks}

Compared to traditional vision tasks under closed-label systems, visual-language tasks are more open and can more comprehensively examine a model's cross-modal understanding and reasoning capabilities. To systematically evaluate the multimodal modeling performance of FUSAR-KLIP, we established a visual-language task benchmark covering three representative tasks: image captioning, image-text retrieval, and visual question answering (VQA). These three tasks correspond to semantic generation, cross-modal matching, and cross-modal reasoning, respectively, and can test the model's generalization performance and multi-dimensional expressive capabilities in remote sensing language understanding from different dimensions.

The experimental configuration is based on the fine-tuning settings of mainstream multimodal models, such as BLIP\cite{li2022blip}, used in the natural image domain, with adjustments made to suit the characteristics of the data. Table \ref{tab7} summarizes the core training parameters for the three visual language tasks. Considering task-specific differences, the captioning task is configured with a larger maximum token length, while the VQA task uses a slightly higher learning rate to accelerate convergence.

\begin{table}[!htb]
	\centering
	\caption{Experimental setup for the visual-language task.}
	\small
	\setlength{\tabcolsep}{5pt}
	\resizebox{0.5\textwidth}{!}{
		\renewcommand{\arraystretch}{1.3}
		\begin{tabular}{l c c c c c c c }
			\hline
			\textbf{Task} & \textbf{Loss Functions} & \textbf{Token} & \textbf{Epoch} & \textbf{LR} & \textbf{Optimizer} &  \\  \hline
			\textbf{Retrieval} & ITC, ITM                & 35                   & 12               & 0.00001               & AdamW                                 \\ 
			\textbf{Caption}     & ITC, ITM, MLM           & 100                  & 12               & 0.00001               & AdamW                                 \\ 
			\textbf{VQA} & ITC, ITM, MLM           & 35                   & 12               & 0.00002               & AdamW                                \\ \hline
		\end{tabular}
	}
	\label{tab7}
\end{table}

In terms of fine-tuning data construction, we generate accurate text based on image detection and segmentation annotations and split the data into training and test sets at a 4:1 ratio. For the captioning and retrieval tasks, the text content includes key information such as the category, quantity, and spatial location of the target. To enhance the diversity of semantic expression, five language templates are designed. For the VQA task, we create question-answer pairs focused on target attributes and ground feature distribution, covering questions related to target counting, type recognition, location positioning, and landform classification. \figref{fig12} presents examples of how the three types of tasks are constructed.

\begin{figure}[!htbp]
	\centerline{
		\phantomsection
		\includegraphics[width=0.35\textwidth]{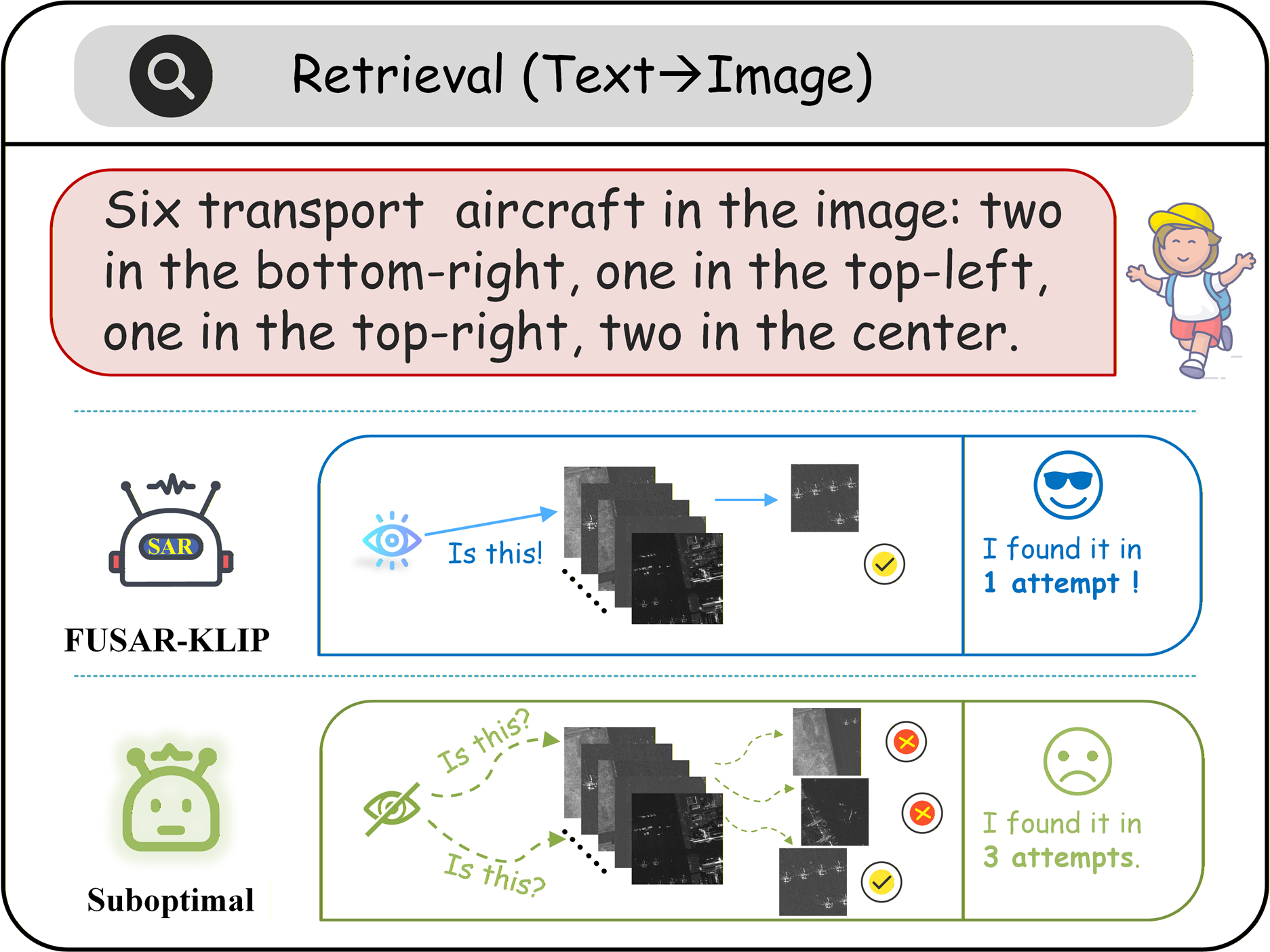}}	
	\caption{Comparison of capabilities from text to image retrieval.}
	\label{fig13}
\end{figure}

\textbf{\textit{1) Image-Text Retrieval:}}
This task evaluates the model’s cross-modal retrieval capability between SAR images and text in both directions: Image→Text and Text→Image, which is crucial for remote sensing image retrieval. Standard $Recall@K$ metrics ($R@1$, $R@5$, $R@10$) are used to assess performance, with $txt\_{r1/r5/r10}$ and $img\_r1/r5/r10$ denoting the Top-K recall rates for text-to-image and image-to-text retrieval, respectively. The average of these six values serves as the overall retrieval metric. Quantitative and qualitative results are shown in Table~\ref{tab8} and \figref{fig13}. As shown in \figref{fig13}, we compare the proposed method with a suboptimal model. FUSAR-KLIP has the highest matching degree for the correct option and can therefore retrieve the corresponding image more accurately.

\textbf{\textit{2) Image Captioning:}}
The Caption task aims to generate accurate descriptions of key targets and their spatial distribution in SAR images using natural language. We evaluate the quality of the image descriptions using four mainstream metrics\cite{li2022blip}: BLEU-4, METEOR, CIDEr, and SPICE. These metrics assess the description quality from four dimensions: phrase matching, semantic consistency, content relevance, and structural integrity. Table \ref{tab8} presents the quantitative evaluation results for each pre-trained model on the Caption task. FUSAR-KLIP achieves leading performance across many indicators, particularly demonstrating significant advantages in CIDEr.

\begin{figure*}[!hbp]
	\centerline{
		\phantomsection
		\includegraphics[width=0.79\textwidth]{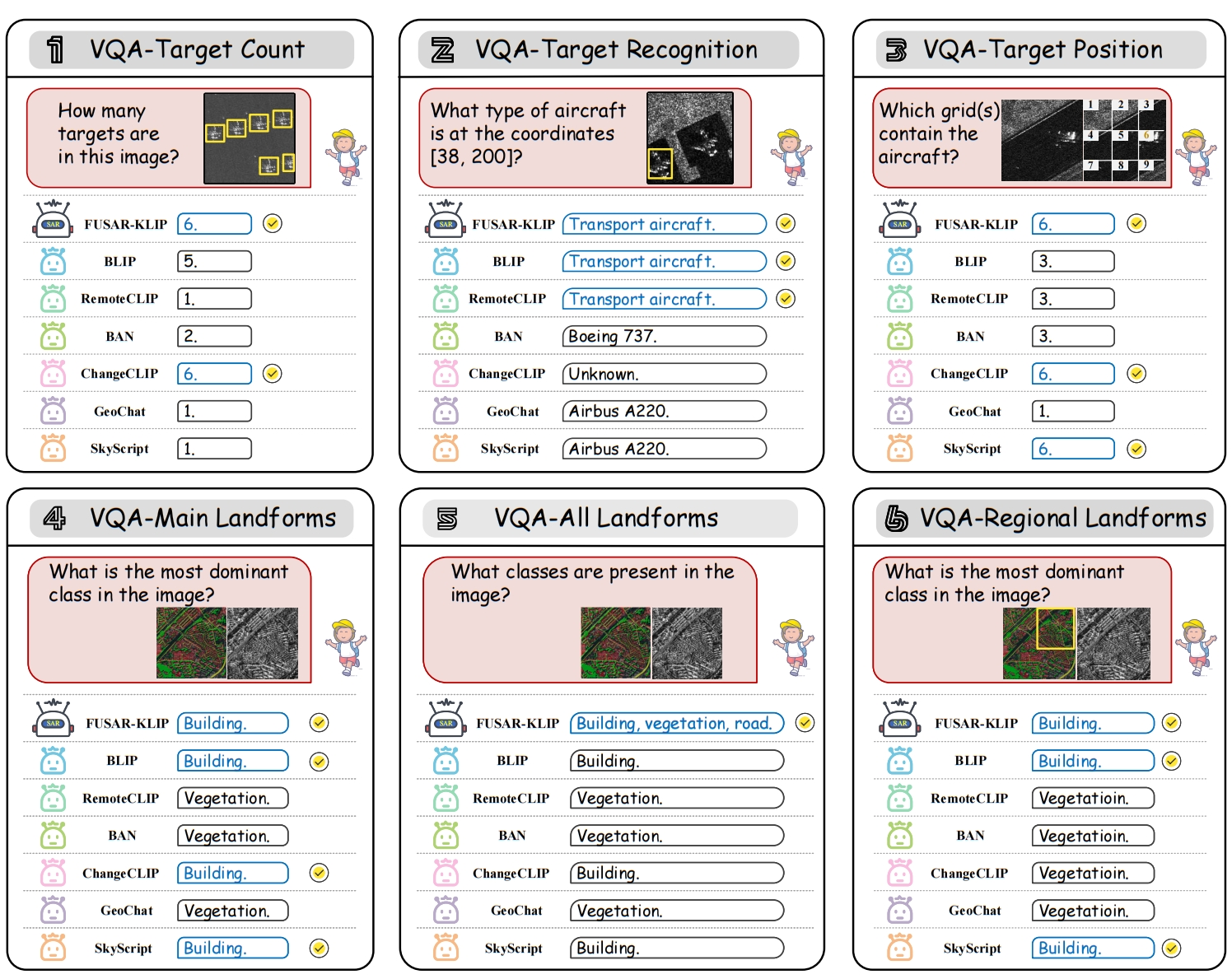}}	
	\caption{The multimodal model's responses to the VQA task, with the correct responses shown \textcolor{blue}{ in color}. Overall, FUSAR-KLIP has the best performance.}
	\label{fig14}
\end{figure*}

\textbf{\textit{3) Visual Question Answering:}}
The VQA task is designed to assess the model’s capability to comprehend SAR image content when presented with natural language queries. Leveraging target detection annotations, we construct three question types: target count estimation, target category recognition, and target location reasoning. In particular, the location reasoning task requires the model to identify the region number containing the target, offering a quantitative evaluation of its spatial understanding. Additionally, based on ground object segmentation and annotation data, we design three landform recognition tasks: identifying the dominant landform type, recognizing the landform type in a specified region, and detecting all landform types present in the image.

Table~\ref{tab8} summarizes the performance across all six tasks, while \figref{fig14} presents qualitative examples of model predictions for representative questions. Experimental results show that FUSAR-KLIP consistently achieves strong performance, especially in target location reasoning and comprehensive landform identification, highlighting its robust reasoning ability in SAR-based visual question answering scenarios.

\subsection{Comparison with general multimodal large models}
\begin{table}[!bp]
	\centering

	\caption{Comparison with General multimodal LLM. Target Rec: Target Recognition. Target Pos: Target Position. Main Land:  Main Landforms. All Land:  All Landforms. Reg Land:  Regional Landforms.}
	\setlength{\tabcolsep}{6pt}
	\resizebox{0.48\textwidth}{!}{  
		\renewcommand{\arraystretch}{1.1}
		\begin{tabular}{ l c c c c c c }
			\hline
			\addlinespace[3pt] 
			\textbf{LLMs} & \textbf{\makecell{Target \\ Count}} & \textbf{\makecell{Target \\ Rec}} & \textbf{\makecell{Target \\Pos}} & \textbf{\makecell{Main \\ Land}} & \textbf{\makecell{ALL \\ Land}} & \textbf{\makecell{Reg \\ Land}} \\ \hline
			GPT-5\cite{OpenAIChatGPT} & 35 & 67 & 28 & 52 & 49 & 47 \\ 
			Gemini-3\cite{google_gemini_web} & 21 & 40 & 41 & 56 & 71 & 42 \\ 
			Grok-4\cite{xai_grok4} & 19 & 29 & 54 & 36 & 67 & 48 \\ 
			Qwen3-VL\cite{bai2025qwen3vltechnicalreport} & 38 & 51 & 27 & 54 & 34 & 40 \\ 
			\rowcolor{lightgray}\textbf{FUSAR-KLIP} & \textbf{98} & \textbf{96} & \textbf{97} & \textbf{99} & \textbf{89} & \textbf{93} \\ \hline
		\end{tabular}
	}
	\label{tab10}
\end{table}

\begin{table*}[!b]
	\centering

	\caption{Benchmarks for visual language tasks, including image-text retrieval, image captioning, and six question-answering tasks. Target Pos: Target Position. Main Land:  Main Landforms. All Land:  All Landforms. Reg Land:  Regional Landforms.  Suboptimal results are \underline{underlined.}}
	\setlength{\tabcolsep}{4pt}
	\resizebox{1.0\textwidth}{!}{
		\renewcommand{\arraystretch}{1.4}
		\begin{tabular}{l c|c c c c c c c|c c c c|c c c c c c }
			\hline
			\multicolumn{2}{c|}{\textbf{Task}} & \multicolumn{7}{c|}{\textbf{Image-Text Retrieval}}&\multicolumn{4}{c|}{\textbf{Image Caption}}&\multicolumn{5}{c}{\textbf{Visual Question Answer}}\\\hline
			\multirow{2}{*}{\textbf{Pretrain Model}} & \multirow{2}{*}{\textbf{Backbone}}& \textbf{TXT} & \textbf{TXT} & \textbf{TXT} & \textbf{IMG} & \textbf{IMG} & \textbf{IMG} & \multirow{2}{*}{\textbf{R}} & \textbf{Bleu} & \textbf{MET-} & \textbf{CID-} & \textbf{SPI-} & \textbf{\makecell{Target}} & \textbf{\makecell{Target}} & \textbf{\makecell{Target }} & \textbf{\makecell{Main }} & \textbf{\makecell{ALL}} & \textbf{\makecell{Reg}}  \\
			&                  & \textbf{\_R1}   & \textbf{\_R5}   & \textbf{\_R10}   & \textbf{\_R1}   & \textbf{\_R5}   & \textbf{\_R10}   &   &\textbf{\_4}  & \textbf{EOR} & \textbf{Er} & \textbf{CE}  & \textbf{Count}  &\textbf{Rec}  &\textbf{Pos}  &\textbf{Land} &\textbf{Land} &\textbf{Land} \\  \hline
			
			RemoteClip                       & ViT-Base                 & 8.07   & 14.91   & 21.12   & 5.34   & 25.34   & 37.76   & 18.76  & 73.53 & 50.09 & 76.24 & 49.45  & 55.48 & 72.49 & 28.68 & 85.79 & 38.80 & 76.09 \\  
			GeoRSClip                        & ViT-Base                 & 9.32   & 16.77   & 25.47   & 9.32   & 26.34   & 43.48   & 21.78  & 73.88 & 49.18 & 71.88 & 48.52  & 55.70 & 77.01 & 16.58 & 84.70 & 44.81 & 73.36 \\ 
			BAN                               & ViT-Base                 & 9.32   & 19.88   & 24.84   & 9.69   & 26.83   & 37.76   & 21.39  & 76.98 & 50.75 & 93.67 & 50.72  & 53.96 & 62.91 & 17.76 & 82.51 & 44.26 & 76.50 \\  
			ChangeClip                        & ViT-Base                 & 10.56  & 18.01  & 28.57  & 7.08  & 26.83   & 41.49   & 22.09  & 72.95 & 50.89 & 67.65 & 48.32  & 55.70 & 22.11 & 22.24 & 85.79 & \underline{47.54} & \underline{77.60} \\  
			Prithvi                            & ViT-Base                 & 6.21   & 11.80   & 14.29   & 2.61   & 13.66   & 22.98   & 11.93  & \underline{77.62} & 50.54 & 73.66 & \underline{51.16}  & 52.01 & 50.50 & 8.55 & 78.69 & 17.49 & 73.50 \\  
			SkyScript                         & ViT-Base                 & 8.70   & 22.36   & 32.92   & 10.31  & 34.78   & 49.94   & 26.50  & 72.68 & 48.71 & 67.92 & 48.20   & 54.18 & 75.81 & 14.47 & 86.34 & 20.77 & 76.64 \\  
			CLIP                              & ViT-Base                 & 14.29  & 19.88   & 23.60   & 9.44   & 24.84   & 34.53   & 21.10  & 75.05 & 51.80 & 67.79 & 49.06  & 56.13 & 69.99 & 26.58 & \underline{86.37} & 42.08 & 76.37 \\  
			BLIP                              & ViT-Base                 & \underline{15.53}  & \underline{29.19}   & \underline{39.75}   & \underline{15.53}  & \underline{52.05}   & \underline{65.96}   & \underline{36.34}  & 76.52 & \underline{56.28} & \underline{104.36} & 49.55  & \underline{76.66} & \underline{89.55} & \underline{79.76} & 82.51 & 39.34 & 77.19 \\
            SARCLIP& ViT-Base&6.21&13.66&22.36 &7.95 &26.21 &37.27 &18.94 & 74.81&50.02 &76.85 &50.23 &54.07 &74.06 &19.74 &84.15 &30.05  &77.59 \\ 
            \hline
			\rowcolor{lightgray}\textbf{FUSAR-KLIP}               & \textbf{ViT-Base}        & \textbf{20.50}  & \textbf{38.51}  & \textbf{50.93}  & \textbf{20.25}  & \textbf{49.57}  & \textbf{68.45}  & \textbf{41.37}  & \textbf{80.01}  & \textbf{57.68}  & \textbf{113.46} & \textbf{51.40} & \textbf{98.70} & \textbf{96.43} & \textbf{97.89} & \textbf{99.45} & \textbf{89.07} & \textbf{93.58} \\ \hline
			RemoteClip                       & ViT-Large                & \underline{11.80}  & 18.01   & 24.22   & 8.70   & \underline{33.04}   & \underline{45.84}   & 23.60  & 76.96 & 50.18 & 80.63 & \underline{51.01}  & \underline{55.27} & \underline{87.72} & \underline{29.72} & 82.51 & 43.72 & 70.90 \\ 
			BAN                               & ViT-Large                & 9.32   & 18.63   & 24.23   & \underline{12.17}  & 29.94   & 44.84   & 23.19  & 72.88 & 50.56 & 4.66  & 47.28  & 53.42 & 60.40 & 16.32 & 86.89 & 9.84 & 75.82
			\\  
			VHM                               & ViT-Large                & 9.32   & 13.66   & 17.39   & 7.45   & 26.34   & 38.26   & 18.74  & 76.96 & 49.96 & 79.07 & 50.67  & 52.23 & 62.34 & 8.68 & 86.88 & 30.05 & 75.68 \\  
			GeoChat                           & ViT-Large                & 5.59   & 11.80   & 21.12   & 6.21   & 21.99   & 35.90   & 17.10  & 76.47 & 50.14 & \underline{83.48} & 49.41  & 52.66 & 73.31 & 8.68 & 79.78 & 30.05 & 77.46
			\\  
			SkyScript                         & ViT-Large                & 8.07   & 14.91   & 23.60   & 10.31  & 26.83   & 42.36   & 21.01  & 77.26 & 50.28 & 76.79 & 50.95  & 52.77 & 85.65 & 18.34 & 84.70 & 43.72 & 78.55 \\  
			BITA                              & ViT-Large                & 6.83   & 11.18   & 18.01   & 6.71   & 23.23   & 33.91   & 16.65  & \underline{77.41} & 50.15 & 80.60 & 50.93  & 52.33 & 78.13 & 8.95 & 86.33 & 19.13 & 77.73 \\  
            SARCLIP& ViT-Large&11.18 &\underline{23.60} &\underline{28.57} &8.57 & 32.04& 43.48& \underline{24.16}& 76.97&\underline{51.01} & 73.34& 50.54& 53.09& 85.96& 9.74& \underline{88.52}& 43.17&\underline{78.83}  \\ 
            \hline
			\rowcolor{lightgray}\textbf{FUSAR-KLIP}               & \textbf{ViT-Large}       & \textbf{26.70}  & \textbf{39.13}  & \textbf{51.55}  & \textbf{29.81}  & \textbf{61.86}  & \textbf{75.90}  & \textbf{47.49}  & \textbf{79.64}  & \textbf{58.66}  & \textbf{117.83} & \textbf{52.11} & \textbf{98.91} & \textbf{93.36} & \textbf{98.16} & \textbf{90.89} & \textbf{96.72} & \textbf{81.42} \\ \hline
		\end{tabular}
	}
	\label{tab8}
\end{table*}

Some commercial multimodal large models with ultra-large parameter volumes (such as GPT-5, Gemini 3, and Grok-4) have demonstrated strong multi-task capabilities. However, due to their closed-source nature, their applicability in the remote sensing field still requires further verification. Given that VQA is the most common application form for multimodal language models, we use the VQA task to assess the reasoning ability of these models on SAR images. The experimental results are shown in Table \ref{tab10}. The general large models perform the weakest in the target counting task, and their accuracy in other tasks is significantly lower compared to the model proposed in this study. This indicates that general multimodal large models suffer from insufficient adaptability in SAR tasks, while the foundational model developed in this research provides robust support for advancing multimodal models in the SAR domain.

\subsection{Ablation experiment}

In the preceding experiments, we systematically validated the overall performance of FUSAR-KLIP across a variety of vision and multimodal tasks. However, the aggregate results do not directly reveal the specific contributions of each design component. To more clearly analyze the roles of data design and model mechanisms in performance improvement, this section further conducts ablation studies.

In the data validation, we selected the latest large-scale SAR text dataset, SARLANG\cite{wei2025sarlang}, for comparison. This dataset depends on target detection annotations and fixed templates to generate image-text pairs, ensuring high semantic accuracy without the need for additional screening or reconstruction. However, as analyzed in \figref{fig9b}, SARLANG exhibits issues such as sparse semantics and limited knowledge coverage.

\begin{figure}[bp]
	\centerline{
		\phantomsection
		\includegraphics[width=0.45\textwidth]{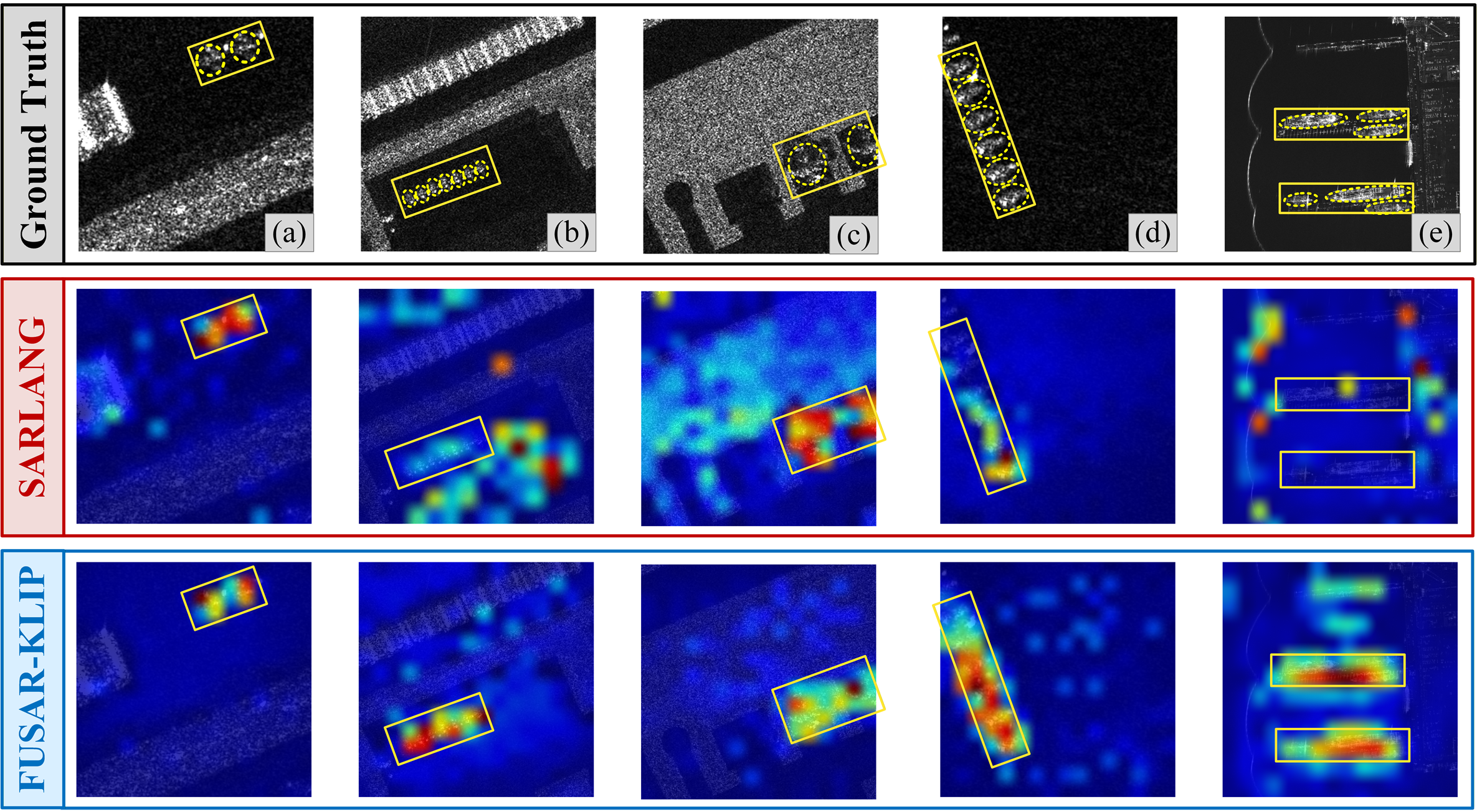}}	
	\caption{Visualization of model feature heatmaps after training on different datasets. FUSAR-KLIP can more accurately extract target features and distinguish background information.}
	\label{fig15}
\end{figure}

To validate the effectiveness of our proposed SCIO strategy, we design a structured evaluation around the screen–filter–reconstruct mechanism, which aims to eliminate low-quality samples and refine textual content to enhance semantic quality and training robustness. Specifically, we conduct two ablation experiments: (1) removing the entire screen–filter–reconstruct pipeline and training directly on the original data; and (2) applying only the screen–filter stage to retain high-quality samples without performing textual reconstruction.

On typical VQA tasks, the experimental results for different configurations are shown in Table \ref{tab11}. The experiments demonstrate that when trained with the FUSAR-GEOVL dataset we constructed, the performance of each task significantly outperforms SARLANG, validating the advanced nature of the proposed data construction method in terms of semantic richness and knowledge guidance. Additionally, both the filtering and reconstruction processes play a critical role in improving performance.

\begin{table}[htbp]
	\Large
	\caption{Ablation Experiments on Data and Modules. }
	\setlength{\tabcolsep}{4pt}
	\resizebox{0.48\textwidth}{!}{  
		\renewcommand{\arraystretch}{1.5}
		\centering
		\begin{tabular}{ l c c | c c c c c c }
			\hline
			\addlinespace[2pt] 
			\textbf{\makecell{Pretrain \\ Dataset}} & \textbf{\makecell{Screen \\ Filter}} & \textbf{Refine} & \textbf{\makecell{Target \\ Count}} & \textbf{\makecell{Target \\ Rec}} & \textbf{\makecell{Target \\Pos}} & \textbf{\makecell{Main \\ Land}} & \textbf{\makecell{ALL \\ Land}} & \textbf{\makecell{Reg \\ Land}} \\ \hline
			SARLANG & \ding{55} & \ding{55} & 71.12 & 86.24 & 92.24 & 94.54 & 83.61 & 78.96 \\ 
			FUSAR-GEOVL & \ding{55} & \ding{55} & 92.83 & 93.86 & 93.68 & 96.17 & 86.34 & 85.52 \\ 
			FUSAR-GEOVL & \ding{51} & \ding{55} & 97.18 & 95.80 & 94.61 & 98.36 & 87.43 & 88.39 \\ 
			\rowcolor{lightgray}\textbf{FUSAR-GEOVL} & \textbf{\ding{51}} & \textbf{\ding{51}} & \textbf{98.70} & \textbf{96.43} & \textbf{97.89} & \textbf{99.45} & \textbf{89.07} & \textbf{93.58} \\ \hline
		\end{tabular}
	}
	\label{tab11}
\end{table}

\figref{fig15} presents a visual comparison of features extracted by different models in the target counting task. From \figref{fig15}[a-b], it is evident that FUSAR-KLIP exhibits stronger resistance to interference. \figref{fig15}[c-e] further shows that the model trained with SARLANG suffers from noticeable target omissions, while FUSAR-KLIP accurately focuses on all target areas.

\section{Conclusion}\label{sec5}

This paper proposes FUSAR-KLIP, the first knowledge-guided multimodal foundational model for SAR, addressing the cognitive gap between general vision and geoscientific interpretation.  Benchmarks across 11 downstream tasks demonstrate that FUSAR-KLIP consistently outperforms 15 mainstream models, particularly in reasoning-heavy tasks like target counting. This result strongly validates that incorporating geographical priors and physical cognition into representation learning is a key path to overcoming the bottlenecks in remote sensing interpretation and bridging the gap between machine representation and human cognition.

More broadly, this work highlights the importance of knowledge-guided modeling in the development of remote sensing foundation models and suggests a unified perspective that integrates geographic priors with cognitive modeling. The proposed framework also offers methodological insights for other remote sensing modalities with strong physical characteristics, such as multispectral and infrared imagery. Future work will focus on expanding data scale and task diversity, as well as further exploring unified modeling across modalities and domains.

\bibliographystyle{IEEEtran}  
\bibliography{reference}  

\end{document}

\end{document}